\documentclass[journal]{IEEEtran}
\usepackage{microtype}    %
\usepackage{amsmath,amssymb}
\usepackage{graphicx}
\usepackage{fancyhdr}
\usepackage{url}
\usepackage{cite}
\usepackage{booktabs} %
\usepackage{paralist} %
\usepackage{longtable}
\usepackage{multirow}
\usepackage[table]{xcolor}
\usepackage{mathtools}
\usepackage{dcolumn} %
\usepackage{multirow}
\usepackage[keeplastbox]{flushend} %
\usepackage{array}
\usepackage{algorithm,algpseudocode}
\usepackage{subfig}
\usepackage{color}
\usepackage[binary-units]{siunitx}
\usepackage{verbatim}
\usepackage[shadow,textsize=footnotesize,color=yellow,colorinlistoftodos]{todonotes}

\newcommand*{\vv}[1]{\vec{\,\mkern0mu#1}}

\algnewcommand{\LineComment}[1]{\State \(\triangleright\) #1}

\graphicspath{{./figs/}}

\title{Relative Pose Based Redundancy Removal:  Collaborative RGB-D Data Transmission in Mobile Visual Sensor Networks}
	
\author{Xiaoqin~Wang$^{1}$\thanks{$^{1}$X. Wang  and T. Drummond are with the ARC Centre of Excellence for Robotic Vision, Monash University, Victoria, 3800, Australia (e-mail: {\scriptsize \textsf{xiaoqin.wang@monash.edu}}, {\scriptsize \textsf{tom.drummond@monash.edu}}).} $\quad$$\quad$ Y. Ahmet \c{S}ekercio\u{g}lu$^{2}$\thanks{$^{2}$Y. A. \c{S}ekercio\u{g}lu, V. Fr\'{e}mont, and E. Natalizio are with the Sorbonne Universit{\'{e}}s, universit{\'{e}} de technologie de Compi{\`{e}}gne, CNRS, UMR 7253 Heudiasyc-CS 60 319, 60 203 Compi{\`{e}}gne Cedex, France (e-mail: {\scriptsize \textsf{asekerci@ieee.org}}, {\scriptsize \textsf{vincent.fremont@utc.fr}}, {\scriptsize \textsf{enrico.natalizio@utc.fr}}).} $\quad$$\quad$   Tom Drummond$^{1}$ \\  Vincent Fr\'{e}mont$^{2}$ $\quad$$\quad$ Enrico Natalizio$^{2}$ $\quad$$\quad$  Isabelle Fantoni$^{3}$\thanks{$^{3}$I. Fantoni is with the Ecole Centrale de Nantes, CNRS, UMR 6004 LS2N, France (e-mail: {\scriptsize \textsf{isabelle.fantoni@ls2n.fr}}).} \thanks{This work was supported by the Australian Research Council   Centre of Excellence for Robotic Vision (project number CE140100016).} \thanks{This work was carried out in the framework of the Labex MS2T and DIVINA challenge team, which were funded by the French Government, through the program \emph{Investments for the Future} managed by the National Agency for Research (Reference ANR-11-IDEX-0004-02).}}%

\begin{document}

\maketitle

\begin{abstract}
  In this paper, the \emph{Relative Pose based Redundancy Removal
    (RPRR)} scheme is presented, which has been designed for mobile
  RGB-D sensor networks operating under bandwidth-constrained
  operational scenarios. The scheme considers a multiview scenario in
  which pairs of sensors observe the same scene from different
  viewpoints, and detect the redundant visual and depth information to
  prevent their transmission leading to a significant improvement in
  wireless channel usage efficiency and power savings.

  We envisage applications in which the environment is static, and
  rapid 3-D mapping of an enclosed area of interest is required, such
  as disaster recovery and support operations after earthquakes or
  industrial accidents.
  
  Experimental results show that wireless channel utilization is
  improved by 250\% and battery consumption is halved when the RPRR
  scheme is used instead of sending the sensor images independently.
\end{abstract}

{\small \textbf{\emph{Note to Practitioners}}--- \textbf{The invention
    of low-cost RGB-D cameras has made large-scale, high-resolution
    3-D sensing for mapping, immersive telepresence, surveillance, or
    environmental sensing tasks easily achievable. A network of mobile
    robots, equipped with the RGB-D cameras can work collaboratively
    and autonomously to rapidly obtain the 3-D map of an area of
    interest. Especially in the cases of natural or industrial
    disasters such as earthquakes, nuclear reactor accidents, or
    building fires, knowing what is happening in a timely manner can
    be critically important, and using a network of cooperating camera
    equipped robots as ``mobile visual sensor networks'' will
    undoubtedly reduce the mapping time while enhancing the overall
    reliability of the operation through redundancy.}

  \textbf{However, the volume of visual and depth data generated by
    these robots is large, which presents a challenge for efficient
    data transmission and storage, particularly over the shared
    wireless channels. The problem is further exacerbated by the
    operation of the robots in such hostile environments leading to
    communication difficulties. In this paper, we present a scheme to
    alleviate this problem. Our scheme allows pairs of visual sensors
    to detect redundant visual information and prevents its
    transmission, gathered when multiple cameras have overlapping
    fields-of-view.}

\begin{sloppypar}
  \textbf{The scheme is computationally lightweight, and so can be
    implemented on battery operated embedded systems without any
    difficulty. Additionally, we also significantly extend the battery
    life of the robots by reducing the transmission load.  }
\end{sloppypar}
}
\vspace*{1ex}

\begin{IEEEkeywords}
  Robotic vision, collaborative coding, relative pose estimation, RGB-D camera,
  visual sensor network, robot networks
\end{IEEEkeywords}

\section{Introduction}

Visual sensor networks (VSNs) allow the capture, processing, and
transmission of per-pixel color information from a variety of
viewpoints.  The inclusion of low-cost compact RGB-D sensors, such as
Microsoft Kinect \cite{kinect}, Asus Xtion \cite{xtion} and Intel
RealSense ZR300 \cite{realsense}, makes VSNs able to collect depth
data as well.

\begin{figure}[h!]
  \centering\includegraphics[width=90mm]{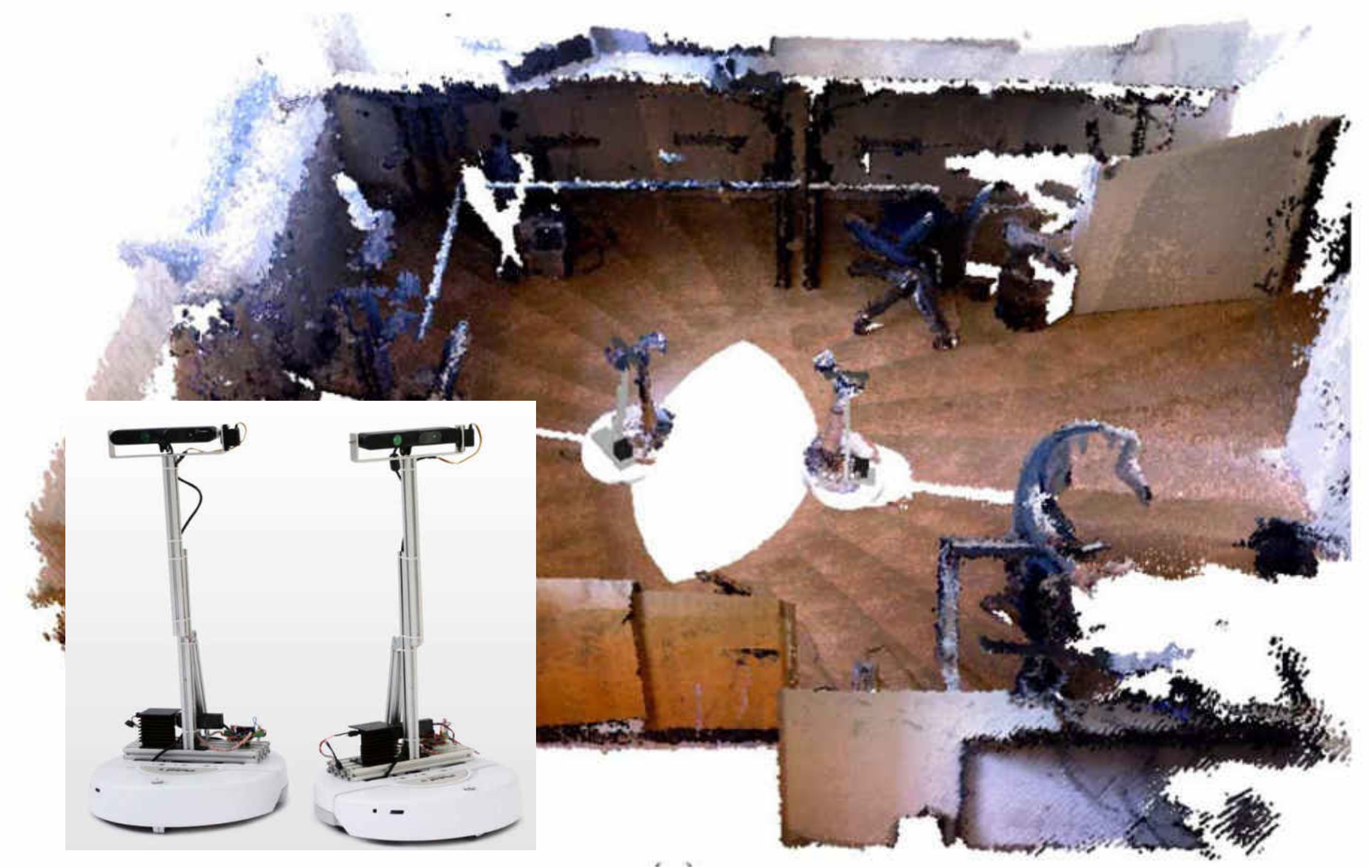}
  \caption{An example of 3-D indoor mapping with two
      simultaneously operating mobile RGB-D sensor platforms
      \cite{Mohanarajah2015}.}
  \label{fig:cloud-based-3d-mapping}
\end{figure}

RGB-D sensor-equipped VSNs can significantly enhance the performance
of conventional applications such as immersive telepresence or mapping
\cite{beck2013immersive,Henry2012,Lemkens2013,Mohanarajah2015},
environment surveillance \cite{Choi2011,Liu2012}, or object
recognition and tracking \cite{Almazan2014,Alexiadis2013,Jing2012} as
well as opening the possibilities for new and innovative applications
like hand gesture recognition \cite{Chong2015}, indoor positioning
systems \cite{Domingo2017} and indoor relocalization
\cite{Li2017}. The value of VSN applications becomes even more
important, especially in places inaccessible to humans, such as
supporting search and rescue operations after earthquakes, industrial
or nuclear accidents. Indeed, examples of mapping (especially indoors)
with networked mobile RGB-D sensors have started to appear in the
research literature (Fig. \ref{fig:cloud-based-3d-mapping}).

RGB-D sensors generate visual and depth data inevitably in huge
quantities. The data volume will be even larger when multiple camera
sensors observe the same scene from different viewpoints and
exchange/gather their measurements to better understand the
environment. As the sensors will most likely be communicating in ad
hoc networking configurations, communication bandwidth will be at a
premium, and will be error-prone and not suitable for continuous data
delivery in large quantities. Moreover, wireless transceivers consume
a significant portion of the available battery power
\cite{aziz2013survey}, and capacity limitation of on-board power
sources should also be considered. Consequently, transmission of
visual and depth information in resource-constrained VSN nodes must be
carefully controlled and minimized as much as possible.

As the same scenery may be observed by multiple sensors (like the
example shown in Fig.~\ref{fig:cloud-based-3d-mapping}), collected
images will inevitably contain a significant amount of correlated
information, and transmission load will be unnecessarily high if all
the captured data are sent. In this paper we focus on this issue, and
present a novel approach to the development of a comprehensive
solution for minimizing the transmission of redundant RGB-D data in
VSNs. Our framework, called \emph{Relative Pose based Redundancy
  Removal (RPRR)}, efficiently removes the redundant information
captured by each sensor before transmission. We designed the RPRR
framework particularly for RGB-D sensor-equipped VSNs which eventually
will need to work in situations with severely limited communication
bandwidth. The scheme operates fully on-board.
In the RPRR framework, the characteristics of depth images, captured
simultaneously with color data, are used to achieve the desired
efficiency. Instead of using a centralized image registration
technique \cite{Lu2007,Merkle2007}, which requires one node to have
full knowledge of the images captured by the others to determine the
correlations, we propose a new approach based on relative pose
estimation between pairs of RGB-D sensors and 3-D
image warping technique \cite{Fehn2003}. The method we propose locally
determines the color and depth information which can only be seen by
one sensor but not the others. Consequently, each sensor is required
to transmit only the uncorrelated information to the remote
station. In order to further reduce the amount of information before
transmission, we apply a conventional coding scheme based on the
discrete wavelet transform \cite{Akansu1992} with progressive coding
features for color images, and a novel lossless differential entropy
coding scheme for depth images (this algorithm was published
in an earlier paper \cite{maxwang2016}). In addition, at the remote monitoring
station, to deal with the artifacts that could occur in the
reconstructed images due to the undersampling problem \cite{Mark1999},
we use our post-processing algorithms. 

Early results of this work were 
presented in \cite{Wang2016}, and in this paper we 
\begin{enumerate}[i.]
\item Add detailed theoretical refinements, practical implementation and
  experimental performance evaluation of the cooperative relative pose
  estimation algorithm \cite{Wang2013}
  (Section~\ref{sec:rel-pos-estimation}),
\item Extend the theoretical development and practical implementation
  of the RPRR scheme for minimizing the transmission of redundant
  RGB-D data collected over multiple sensors with large pose
  differences (Section~\ref{sec:3D-warping}),
\item Describe the lightweight crack and ghost artifacts
  removal algorithms as a solution to the undersampling problem
  (Section~\ref{sec:filling}), and
\item Include detailed experimental evaluation of
  wireless channel capacity utilization and energy consumption
  (Section~\ref{sec:bandwidth-energy}).
\end{enumerate}
In the following sections of the paper, after a discussion of the
related work, we present the details of the RPRR framework in
Section \ref{sec:overview}, experimental results and their
analysis can be found in Section \ref{sec:results}, followed by our
concluding remarks.

\section{Related Work}

A number of solutions exists in the research literature that
intend to remove or minimize the correlated data for transmission in
VSNs. They can be broadly classified into three groups:
\begin{inparaenum}[(i)]
\item Optimal camera selection,
\item Collaborative compression and transmission, and
\item Distributed source coding.
\end{inparaenum}
The optimal camera selection algorithms
\cite{Chow2007,Bai2009,Bai2010,RuiDai2009,Colonnese2013} attempt to
group the camera sensors with overlapping fields-of-view (FoVs) into
clusters and only activate the sensor which can capture the image with
the highest number of feature points. The pioneering work presented in
\cite{RuiDai2009} demonstrated that a correlation-based algorithm can
be designed for selecting a suitable group of cameras communicating
toward a sink so that the amount of information from the selected
cameras can be maximized. Based on this work, in \cite{Colonnese2013},
the concept of ``common sensed area'' was proposed between two
views to measure the efficiency of multiview video coding techniques
and reduce the amount of information transmitted in VSNs. These
algorithms operate under the assumption that the images captured by a
small number of camera sensors in one cluster are good enough to
represent the information of the scene/object. In these approaches,
the location and orientation of the camera sensors are used to
establish clusters, and a variety of existing feature detection
algorithms \cite{Bay2008,Drummond2006} or place recognition approaches \cite{Li2017,Lowry2016} are used to determine the
similarity between captured images in each cluster. However, the
occlusions in FoVs may cause significant differences between the
images captured by cameras with very similar sensing
directions. Therefore, the assumption is not realistic and this kind
of approach is not applicable in many situations.

The collaborative compression and transmission methods
\cite{Chia2009,Chia2012,Cen2017,Wu2007,Shaheen2015} jointly encode the
captured multi-view images. The spatial correlation is explored and
removed at encoders by image registration algorithms.  Only the
uncorrelated visual content is delivered in the network after being jointly
encoded by some recent coding techniques (e.g., Multiview Video Coding
(MVC) \cite{Vetro2011,Redondi2016}) and compressive sensing approaches \cite{Ebrahim2015,Zhang2017}. However, at least one node in the network is
required to have the full set of images captured by the other sensors
in order to perform image registration. This means that the redundant
information cannot be removed completely and still needs to be
transmitted at least once. Moreover, as color images do not
contain a full 3-D representation of a scene, these methods introduce
distortions and errors when the relative poses (location and
orientation) between sensors are not pure rotation or translation, or
the scenes have complex geometrical structures and occlusions.

Distributed source coding (DSC) algorithms
\cite{Deli2012,yeo2010,Heng2017,Hanca2016,Luong2016} are other
promising approaches that can be used to reduce the redundant data in
multiview VSN scenarios. Each DSC encoder operates independently, but
at the same time, relies on joint decoding operations at the sink
(remote monitoring station). The advantage of these approaches is that
the camera sensors do not need to directly communicate the captured visual
information with others in the network. Furthermore, these
algorithms shift the computational complexity from the sensor nodes to
the remote monitoring station, which fits the needs of VSNs
well. However, the side information must be
predicted as accurately as possible and the correlation structure
should be able to be identified at the decoder side (remote monitoring
station), without an accurate knowledge of the network topology and the
poses of the sensors. These are the main disadvantages that prevent
DSC algorithms from being widely implemented.  A detailed discussion
on multi-view image compression and transmission schemes in VSNs is
presented in \cite{Wang2013:survey}. 

The algorithms mentioned above focus only on color (RGB) data.  Only a
few studies have been reported
\cite{Kadkhodamohammadi2017,Ju2013,Mohanarajah2015} that use RGB-D sensors in
VSNs, as their use in networked robotics scenarios has not yet become
widespread. Consequently, our extensive review of the research
literature has not identified any earlier studies that attempt to
develop an efficient coding system that aims to maximize the bandwidth
usage and minimize the energy consumption for RGB-D equipped VSNs.

\section{Relative Pose Based Redundancy Removal (RPRR) Framework}
\label{sec:overview}

\subsection{Overview}

\begin{figure}%
  \centering\includegraphics[width=70mm]{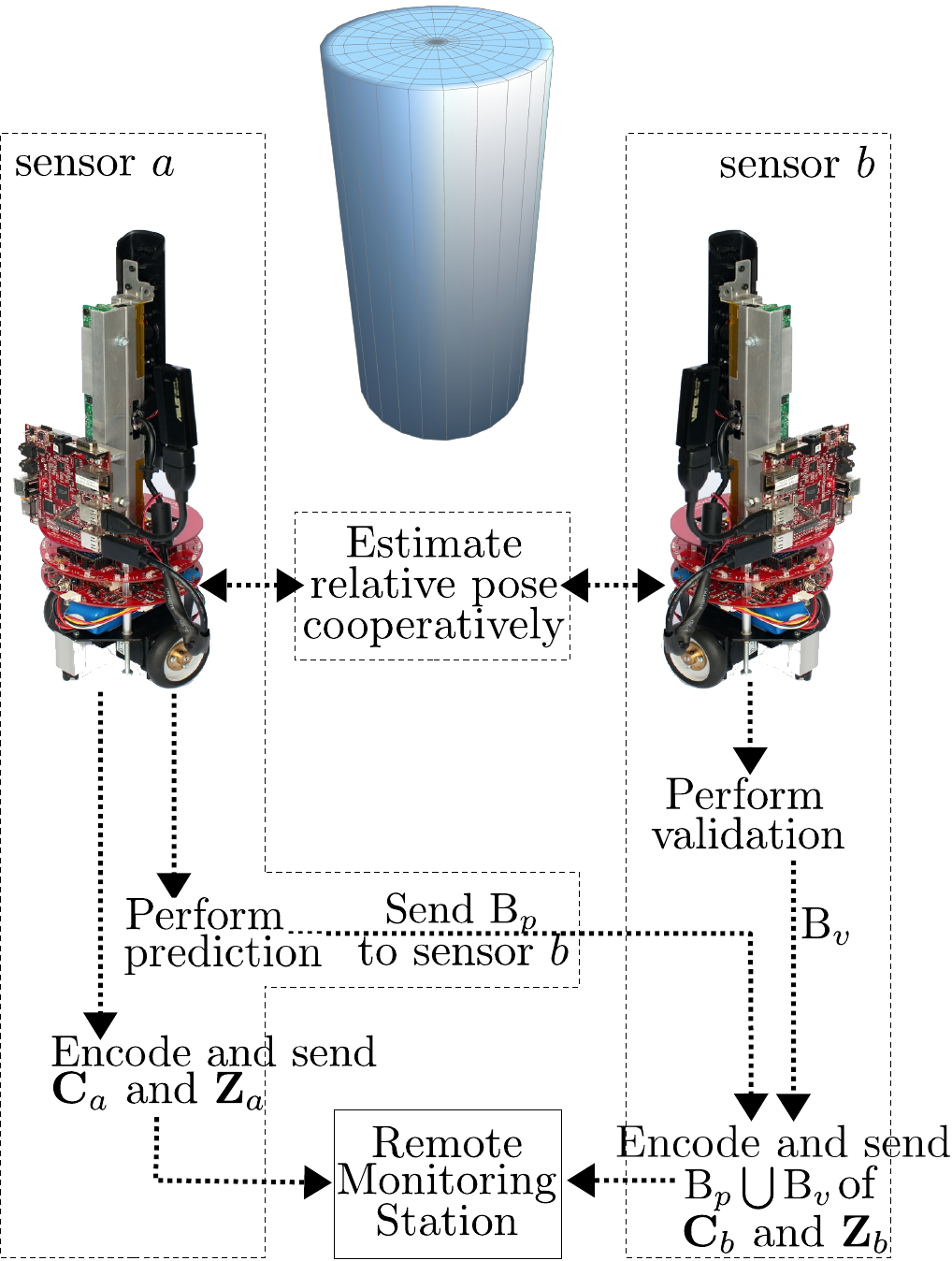}
  \caption{Operational overview of the RPRR framework. The sensors
    first cooperatively estimate their relative poses by using the
    algorithm shown in Fig. \ref{fig:rel-pose-estimation}, then
    identify the non-overlapping image blocks to send only the
    non-redundant visual information to the remote monitoring
    station.}
  \label{fig:overview}
\end{figure}

In a mobile VSN tasked with mapping a region using RGB-D sensors, it
is highly possible that multiple sensors will observe the same scene
from different viewpoints. Because of this, scenery captured by the
sensors with overlapping FoVs will have a significant level of
correlated and redundant information. Here, our goal is to efficiently
extract and encode the uncorrelated RGB-D information, and avoid
transmitting the same surface geometry and color information
repeatedly.

Consider the two sensors, $a$ and $b$, of this VSN with overlapping
FoVs. Let $\mathbf{Z}_{a}$ and $\mathbf{Z}_{b}$ denote a pair of depth
images returned by these sensors, and $\mathbf{C}_{a}$ and
$\mathbf{C}_{b}$ are the corresponding color images. In the encoding
procedure, we first estimate the location and orientation of one
sensor relative to the other. Then, correlated and redundant
information in color and depth images are identified to minimize
unnecessary data transmissions to the central monitoring station. To
achieve this, by using the relative pose information, sensor $a$
computes a prediction of $\mathbf{Z}_{b}$ to determine the depth and
color information exists only in $\mathbf{Z}_{b}$ but not in
$\mathbf{Z}_{a}$. Then, it informs sensor $b$ for it to send only the
uncorrelated depth and corresponding color information in
$\mathbf{Z}_{b}$ and $\mathbf{C}_{b}$. To further improve the wireless
channel capacity usage, depth image data is compressed with our own
\emph{Differential Huffman Coding with Multiple Lookup Tables (DHC-M)}
method \cite{maxwang2016}, and color images are compressed with
Progressive Graphics File (PGF) scheme \cite{stamm2002} prior to their
transmission.

At the remote monitoring station, to improve the image quality, we
apply algorithms for removal of the visual artifacts that may be
introduced during the image reconstruction process.

A high-level view of the operation of the system is shown in
Fig.~\ref{fig:overview}. A detailed explanation of each step is
provided in the following sections.

\subsection{Relative Pose Estimation}
\label{sec:rel-pos-estimation}

As an RGB-D sensor can provide a continuous measurement of the 3-D
structure of the environment, the relative pose between two RGB-D
sensors can be estimated through explicit matching of surface
geometries in the overlapping regions within their FoVs. A variety of
algorithms have been proposed to determine whether multiple cameras
are looking at the same scene, such as vision-based
\cite{Bai2010,San2007}, or geometry-based \cite{Dai2009,Ma2005}
methods. Here, we assume that the sensors use one of these approaches
to detect whether they are observing the same scene. Afterwards, as
explained below, with our relative pose estimation algorithm, the
sensors accurately estimate their relative position and orientation
(relative pose).

The relative pose between the RGB-D sensors $a$ and $b$ can be represented
by a transformation matrix,  
\begin{align*}
  \textbf{M}_{ab} = \begin{bmatrix}
    \  & \textbf{R} & \  & \textbf{t}\\
    0 & 0 & 0 & 1
  \end{bmatrix}
\end{align*}
in SE(3) \cite{corke2011-ch02}, where $\textbf{R}$ is a $3 \times 3$
rotation matrix and $\textbf{t}$ is a $3 \times 1$ translation
vector. The transformation matrix $\mathbf{M_{ab}}$ represents the six
degrees of freedom (6DoF) motion model, which not only describes the
relative pose between two sensors and also the transformation of the
structure between the depth images captured by both sensors.

The transformation matrix $\textbf{M}_{ab}$ can be estimated by
matching the surface geometries captured by two sensors. Taking
advantage of the depth image characteristics, the depth pixels in a
frame captured by sensor $b$ can be mapped to a frame captured by
sensor $a$. Consider the vector $\mathbf{p}_e = [x~ y~ z~ 1]^{T}$
which represents a real world point in Euclidean space by using
homogeneous coordinates. Given the following intrinsic parameters of
an RGB-D sensor:
\begin{inparaenum}[(i)]
\item principal point
coordinates $(i_{c}$, $j_{c})$ and 
\item focal length of the camera
$(f_{x}$, $f_{y})$, 
\end{inparaenum}
$\mathbf p_e$ can be estimated from the
corresponding pixel in a depth image by using the pinhole camera model
as
\begin{equation*}
\frac{1}{z}\mathbf{p}_{e} = \frac{1}{z}
\begin{bmatrix}  x & y & z & 1 \end{bmatrix}^{T} 
               =
               \begin{bmatrix}
       \frac{i - i_{c}}{f_{x}} & \frac{j - j_{c}}{f_{y}} & 1 & \frac{1}{z}
               \end{bmatrix}^{T}
\end{equation*}
where $(i,j)$ denotes the pixel coordinates of the projection of this
real world point in the depth image, and $z$ is the corresponding
depth value reported by the camera.

In the discussion that follows, we assume that $\mathbf{p_e}$ can be
observed by both mobile RGB-D sensors $a$ and $b$, and the projections
of $\mathbf p_e$ are located at pixel coordinates $(i_{a},j_{a})$ and
$(i_b,j_b)$ on the depth images $\textbf{Z}_{a}$ and $\textbf{Z}_{b}$
respectively. Under the assumption that the world coordinate system
is equal to the mobile sensor coordinate system, and the intrinsic
parameters of both sensors are identical, the depth pixel (projection)
at $(i_{a},j_{a})$ in $\textbf{Z}_{a}$ can establish a relationship
between the depth pixel at $(i_b,j_b)$ in $\textbf{Z}_{b}$ as follows,
\begin{equation} 
  \begin{bmatrix}
    \frac{i_{b}-i_{c}}{f_{x}} \!&\! \frac{j_{b}-j_{c}}{f_{y}} \!&\! 1 \!&\! \frac{1}{z_{b}}
  \end{bmatrix}^{T} \!\!=
  \textbf{M}_{ab}\!\!\begin{bmatrix}
    \frac{i_{a} - i_{c}}{f_{x}} \!&\! \frac{j_{a} - j_{c}}{f_{y}} \!&\! 1 \!&\! \frac{1}{z_{a}}
  \end{bmatrix}^{T}
  \label{eq:warp}
\end{equation}
and, to simplify the equation, by doing some rudimentary algebraic
substitutions we obtain 
\begin{align*} 
\left[u_{b} \quad v_{b} \quad 1 \quad q_{b}\right]^{T} =
  \mathbf{M}_{ab}\left[u_{a} \quad v_{a} \quad 1 \quad q_{a}\right]^{T}
\end{align*}
in inverse depth coordinates. 

We now need to estimate $\mathbf{M}_{ab}$. To accomplish an accurate
estimate of $\mathbf{M}_{ab}$ we have developed a Iterative Closest
Point (ICP) algorithm which operates in a distributed fashion by using
the explicit registration of surface geometries extracted from the
depth frames captured by two sensors \cite{Wang2013}. It delivers
robust results especially in circumstances with heavy occlusion. In
our distributed algorithm the registration problem is approached by
iteratively minimizing a cost function whose error metric is defined
based on the bidirectional point-to-plane geometrical relationship as
explained in the following paragraphs.

\begin{figure}%
  \centering\includegraphics{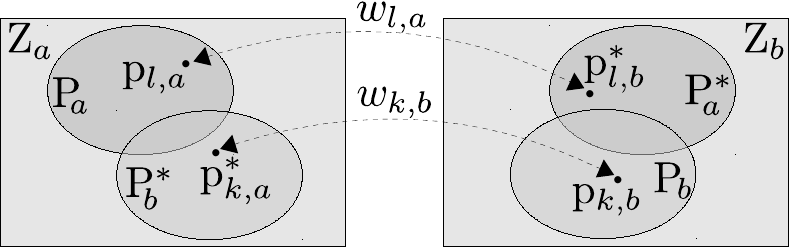}
  \caption{Two sets of points
    ($\mathrm{P}_{\!a} \subset \mathbf{Z}_{a}$ and
    $\mathrm{P}_b \subset \mathbf{Z}_{b}$) sampled from the depth
    images $\textbf{Z}_{a}$ and $\textbf{Z}_{b}$, and their
    corresponding point sets $\mathrm{P}_{\!a}^{*}$ and
    $\mathrm{P}_{\!b}^{*}$. $\mathrm{P}_{\!a}$ and $\mathrm{P}_b$ have
    $N_a$ and $N_b$ number of elements respectively. For finding the
    point sets, project-and-walk method is used with a neighborhood
    size of $7\!\times\!7$ pixels based on the nearest neighbor
    criteria as proposed in \cite{benjemaa1999fast}.}
  \label{fig:correspondences}
\end{figure}

\begin{figure*}
  \centering\includegraphics[width=0.75\linewidth]{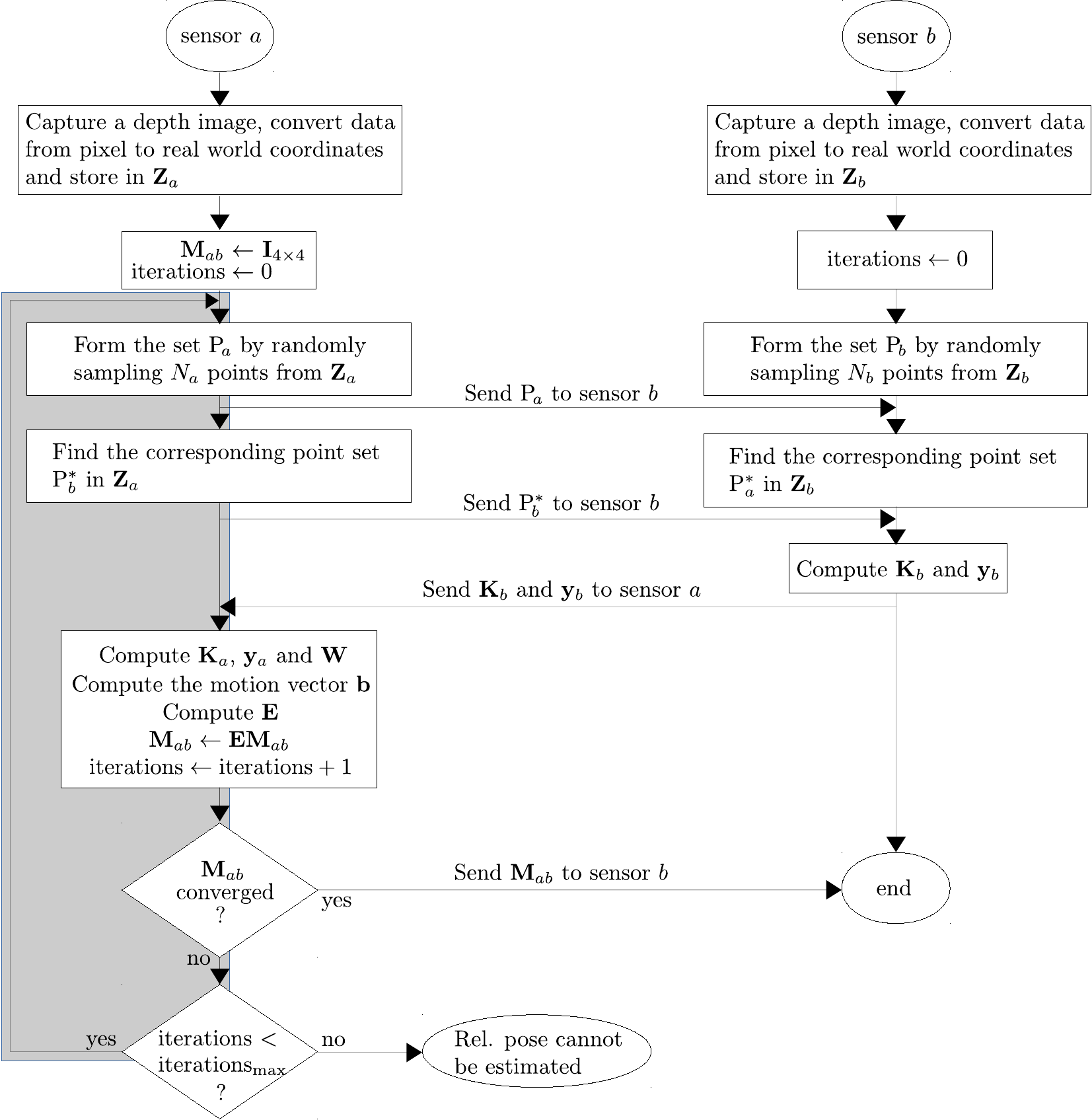}
  \caption{Operation of the cooperative relative pose estimation
    algorithm. The algorithm is distributed over two sensors, and
    operates iteratively (denoted in gray) until it converges or
    maximum number of iterations is reached. We have used the
    convergence criterion presented in \cite{Lui2012}, and
    $\text{iterations}_{\max}$ is set as 50
    (Sec. \ref{sec:relative}).}
  \label{fig:rel-pose-estimation}
\end{figure*}

Let $\mathrm{P}_{\!a} = \{\mathbf{p}_{l,a}, l = 1,2,\dots,N_a\}$ and
$\mathrm{P}_b = \{\mathbf{p}_{k,b}, k = 1,2,\dots,N_b\}$ denote two sets of
measurements sampled from $\mathbf{Z}_a$ and $\mathbf{Z}_b$. Let us
assume that the correspondences for $N = N_{a} + N_{b}$
pairs\footnote{ A typical depth image may have hundreds of thousands
  of points, therefore running algorithms on the full point cloud is
  computationally expensive. In order to alleviate this problem, a
  commonly used method is to subsample the data for speeding up the
  operation with the cost of reduced accuracy. This is a fundamental
  trade-off of ICP performance: Registration by using dense point
  clouds yields a more accurate alignment, however it needs longer
  processing time to complete. On the other hand, a subsampled point
  cloud results in lower accuracy, but requires a significantly
  shorter processing time.  Thus, for the best ICP (and its variants)
  performance, striking a balance between accuracy and processing time
  is required by considering the timing requirements for obtaining
  results and the computational resources available. Considering
  these, after conducting a series of experiments on our sensor
  platforms, we have chosen $N_{a} = N_{b} = 250$.}  of points
$(\mathbf{p}_{l,a} \!\! \leftrightarrow \! \mathbf{p}^{*}_{l,b})$ and
$(\mathbf{p}_{k,b} \!\!\leftrightarrow \!\mathbf{p}^{*}_{k,a})$ are
established to form the sets $\mathrm{P}_{\!a}^*$ and $\mathrm{P}_b^*$.  Here,
$\mathbf{p}^{*}_{l,b} \in \mathrm{P}_{\!a}^*$ (where
$\mathrm{P}_{\!a}^* \subset \mathbf{Z}_{b}$) is the corresponding point of
$\mathbf{p}_{l,a}$, and $\mathbf{p}^{*}_{k,a} \in \mathrm{P}_b^*$ (where
$\mathrm{P}_b^* \subset \mathbf{Z}_{a}$) is the corresponding point of
$\mathbf{p}_{k,b}$ (see Fig.~\ref{fig:correspondences}). Then, the
transformation matrix $\mathbf{M}_{ab}$ can be estimated by minimizing
the bidirectional point-to-plane error metric $\mathcal{C}$, expressed
in normal least squares form as
\begin{multline}
  \mathcal{C} = \sum_{l=1}^{N_{a}} \left[w_{l,a}(\mathbf{M}_{ab}      \mathbf{p}_{l,a}   - \mathbf{p}^{*}_{l,b}) \cdot  \vv{\mathbf{n}}^*_{l,b} \right]^{2} +  \\
                \sum_{k=1}^{N_{b}} \left[w_{k,b}(\mathbf{M}_{ab}^{-1}  \mathbf{p}^{*}_{k,a} - \mathbf{p}_{k,b})     \cdot  \vv{\mathbf{n}}_{k,b} \right]^{2}
  \label{eq:bidirectional}
\end{multline}
where $w_{l,a}$ and $w_{k,b}$ are the weight parameters for the
correspondences established in opposite directions between the pairs,
$(\cdot)$ is the dot product operator,
\begin{align}
  \vv{\mathbf{n}}^*_{l,b} \,=&  \begin{bmatrix} \beta^*_{l,b} & \gamma^*_{l,b}  & \delta^*_{l,b} & 0 \end{bmatrix}^{T} 
  \label{eq:weight}
\end{align}
and
\begin{align}
  \vv{\mathbf{n}}_{k,b}   =&  \begin{bmatrix} \beta_{k,b} & \gamma_{k,b} & \delta_{k,b} & 0 \end{bmatrix}^{T}
  \label{eq:weightt}
\end{align}
are the surface normals at the points $\mathbf{p}^*_{l,b}$ and
$\mathbf{p}_{k,b}$.  The cost function presented in
Eq. \ref{eq:bidirectional} consists of two parts:
\begin{inparaenum}[(i)]
\item the sum of squared distances from $\mathbf{Z}_{a}$ to
  $\mathbf{Z}_{b}$, and
\item the sum of squared distances from $\mathbf{Z}_{b}$ to
  $\mathbf{Z}_{a}$.
\end{inparaenum}
The estimation of $\mathbf{M}_{ab}$ can be done by iteratively
re-weighting the least squares operation in an ICP framework. Based on
this principle, we have created the distributed algorithm which has
two complementary components running concurrently on sensors $a$ and
$b$ as shown in Fig. \ref{fig:rel-pose-estimation}.

On sensor $a$, in the first iteration, $\mathbf{M}_{ab}$ is
initialized as the identity matrix. Afterwards, in this coarse-to-fine
algorithm, by using the information sent by sensor $b$, each iteration
generates an update $\mathbf{E}$ to the sensor's pose which modifies
the transformation matrix $\mathbf{M}_{ab}$. $\mathbf{E}$ takes the
same form as $\mathbf{M}_{ab}$, and can be parameterized by a
6-dimensional motion vector having the elements
$\alpha_{1},\alpha_{2},\dotsc, \alpha_{6}$ via the exponential map and
their corresponding group generator matrices
$\mathbf{G}_{1}, \mathbf{G}_{1}, \dotsc, \mathbf{G}_{6}$ as
\begin{equation}
  \mathbf{E}=\exp\!\!\left(\sum_{j=1}^{6}\alpha_{j}\mathbf{G}_{j}\right)
  \label{eq:MV}
\end{equation}
where 
\begin{equation*}
  \begin{matrix*}[l]
    \mathbf{G}_{1} =
    \begin{bmatrix*}[r]
       0  & 0 & 0 & 1 \\
       0  & 0 & 0 & 0 \\
       0  & 0 & 0 & 0 \\
       0  & 0 & 0 & 0 \\
    \end{bmatrix*} &
    \mathbf{G}_{2} =
    \begin{bmatrix*}[r]
       0  & 0 & 0 & 0 \\
       0  & 0 & 0 & 1 \\
      0  & 0 & 0 & 0 \\
        0  & 0 & 0 & 0 \\
    \end{bmatrix*}
  \end{matrix*}
\end{equation*}
\begin{equation*}
  \begin{matrix*}[l]
    \mathbf{G}_{3} = 
    \begin{bmatrix*}[r]
        0  & 0 & 0 & 0 \\
        0  & 0 & 0 & 0 \\
        0  & 0 & 0 & 1 \\
        0  & 0 & 0 & 0 \\
    \end{bmatrix*} &
    \mathbf{G}_{4} = 
    \begin{bmatrix*}[r]
        0 & 0 & 0 & 0 \\
        0 & 0 & -1 & 0 \\
        0 & 1 & 0 & 0 \\
        0 & 0 & 0 & 0 \\
    \end{bmatrix*} 
  \end{matrix*}
\end{equation*}
\begin{equation*}
  \begin{matrix*}[l]
    \mathbf{G}_{5} = 
    \begin{bmatrix*}[r]
      0 & 0 & 1 & 0 \\
      0 & 0 & 0 & 0 \\
      -1 & 0 & 0 & 0 \\
      0 & 0 & 0 & 0 \\
    \end{bmatrix*} &
    \mathbf{G}_{6} = 
    \begin{bmatrix*}[r]
      0 & -1 & 0 & 0 \\
      1 & 0 & 0 & 0 \\
      0 & 0 & 0 & 0 \\
      0 & 0 & 0 & 0 \\
    \end{bmatrix*} 
  \end{matrix*}
\end{equation*}
Here $\mathbf{G}_{1}$, $\mathbf{G}_{2}$ and $\mathbf{G}_{3}$ are the
generators of translations in $x$, $y$ and $z$ directions, while
$\mathbf{G}_{4}$, $\mathbf{G}_{5}$ and $\mathbf{G}_{6}$ are rotations
about $x$, $y$ and $z$ axes respectively. For details
\cite{drummond02,drummond-3d-geometry} can be referred. The task then
becomes as finding the elements of the six-dimensional motion vector
\begin{equation*}
 \mathbf{b} = \begin{bmatrix} \alpha_{1} & \alpha_{2} & \alpha_{3} & \alpha_{4} & \alpha_{5} &\alpha_{6} \end{bmatrix}^{T}
\end{equation*}
that describe the relative pose. By determining the partial
derivatives of $u_{b}$, $v_{b}$ and $q_{b}$ with respect to the
unknown elements of $\mathbf{b}$, the Jacobian matrix for each
established corresponding point pair can be obtained as
\begin{align}
  \mathbf{J} = \begin{bmatrix}
    q_{a} & 0 & -u_{a}q_{a} & -u_{a}v_{a} &  1+u_{a}^{2} & -v_{a}  \\
    0 & q_{a} & -v_{a}q_{a} & -1-v_{a}^{2} &  v_{a}u_{a} & u_{a}  \\
    0 & 0 & -q_{a}^{2} & -v_{a}q_{a} & u_{a}q_{a} & 0  \\[0.3em]
  \end{bmatrix}
\end{align}
The six-dimensional motion vector $\mathbf{b}$, which minimizes
Eq.~\ref{eq:bidirectional}, is then determined iteratively by the
least squares solution
\begin{align}
  \mathbf{b} =
  (\mathbf{K}^{T}\mathbf{W}\mathbf{K})^{-1}\mathbf{K}^{T}\mathbf{W}\mathbf{y}
  \label{eq:B}
\end{align}
in which %
\begin{equation}
  \mathbf{K} =
  \begin{bmatrix}
    \mathbf{K}_b \\
    \mathbf{K}_a 
  \end{bmatrix}
  \label{eq:K}
\end{equation}
where,
\begin{align}
  \mathbf{K}_b =
  \begin{bmatrix}
    (\vv{\mathbf{n}}_{1,b}^{*\prime})^{T}    \mathbf{J}_{1}^{}  \\
    \vdots \quad \vdots \\
    (\vv{\mathbf{n}}_{l,b}^{*\prime})^{T}    \mathbf{J}_{l}^{}  \\
    \vdots \quad \vdots \\
    (\vv{\mathbf{n}}_{N_{a},b}^{*\prime})^{T} \mathbf{J}_{N_{a}}^{} 
  \end{bmatrix}
&&
  \mathbf{K}_a =
  \begin{bmatrix}
    (\vv{\mathbf{n}}_{1,b}^{\prime})^{T}     \mathbf{J}_{1}^{*} \\
    \vdots \quad \vdots \\
    (\vv{\mathbf{n}}_{k,b}^{\prime})^{T}     \mathbf{J}_{k}^{*} \\
    \vdots \quad \vdots \\
    (\vv{\mathbf{n}}_{N_{b},b}^{\prime})^{T}  \mathbf{J}_{N_{b}}^{*}
  \end{bmatrix}
\end{align}
and
$\vv{\mathbf{n}}_{l,b}^{*\prime} = \begin{bmatrix} \beta_{l,b} &
  \gamma_{l,b} & \delta_{l,b} \end{bmatrix}^{T}$,
$\vv{\mathbf{n}}_{k,b}^{\prime} = \begin{bmatrix} \beta_{k,b} &
  \gamma_{k,b} & \delta_{k,b} \end{bmatrix}^{T}$
are the surface normals at the points
$\mathbf{p}_{l,b}^{*} \in \mathrm{P}_{\!a}^*$ and
$\mathbf{p}_{k,b}^{} \in \mathrm{P}_b^{}$ expressed in a slightly
different form than in Eqs.~\ref{eq:weight} and \ref{eq:weightt}, 
and
\begin{equation}
\mathbf{y} = 
  \begin{bmatrix}
    \mathbf{y}_b \\
    \mathbf{y}_a \\
  \end{bmatrix}
  \label{eq:Y}
\end{equation}
where 
\begin{align}
\mathbf{y}_b = \begin{bmatrix}
    -(\mathbf{p}_{1,a}-\mathbf{p}_{1,b}^{*}) \cdot \vv{\mathbf{n}}_{1,b}^{*} \\
    \vdots   \\
    -(\mathbf{p}_{N_{a},a}^{} -\mathbf{p}_{N_{a},b}^{*}) \cdot \vv{\mathbf{n}}_{N_{a},b}^*  \\
  \end{bmatrix}
\\
\mathbf{y}_a = \begin{bmatrix}
    -(\mathbf{p}_{1,a}^{*}-\mathbf{p}_{1,b}^{}) \cdot \vv{\mathbf{n}}_{1,b}^{}   \\
    \vdots   \\
    -(\mathbf{p}_{N_{b},a}^{*}-\mathbf{p}_{N_{b},b}^{}) \cdot \vv{\mathbf{n}}_{N_{b},b}^{}   \\
  \end{bmatrix} 
\end{align}
and $(\cdot)$ is the dot product operator. Also
\begin{equation}
\mathbf{W} = \begin{bmatrix}
w_{1,1} &\cdots&  0 \\
\vdots & \ddots& \vdots \\
0 & \cdots & w_{\!N,N}^{}
\end{bmatrix}
\end{equation}
contains the weightings for the bidirectional point-to-plane correspondences.
As reported in \cite{Holland77}, different weighting functions lead to
various probability distributions. Based on our experiments, we have
found that the asymmetric weighting function
\begin{align}
w_{a,b} = \left\{
\begin{array}{l l}
c/[c+(z_a-z_b)] & \quad \text{if $z_b \leq z_a$ }\\
c/[c+(z_a-z_b)^{2}] & \quad \text{otherwise}\\
\end{array} \right.
\end{align}
yields satisfactory results. Here $z_a$ and $z_b$ are the depth
values of corresponding points in two depth images, and $c$ is the mean of
differences between the depth values of all corresponding points.

To detect the convergence of our algorithm, we use
the thresholds for the ICP framework presented in \cite{Lui2012}. Once
the algorithm converges, the registration is considered completed
and $\mathbf{M}_{ab}$ is used for the elimination of the redundant 
data in transmissions as explained in Section \ref{sec:3D-warping}.

The sensors exchange very small amounts of information by using this
algorithm making the process very bandwidth-efficient fitting the
requirements of VSNs. We present an in-depth analysis of message
exchange complexity in Section \ref{sec:relative}.

\subsection{Identification of Redundant Regions in Images}
\label{sec:3D-warping}

\subsubsection{Prediction}

Sensor $a$, by using the relative pose information $\textbf{M}_{ab}$,
can now apply Eq.~\ref{eq:warp} on each pixel in $\textbf{Z}_{a}$ to
create a predicted a depth image $\textbf{Z}_{b}^{*}$, which is
virtually captured from sensor $b$'s viewpoint. In this process
though, it could happen that two or more different depth pixels are
warped into the same pixel coordinates in $\textbf{Z}_b^*$. This
over-sampling issue could occur because some 3-D world points are
occluded by the other ones at the new viewpoint. In order to solve
this problem, we always compare the depth values of the pixels warped
to the same coordinates, and the pixel with the closest range
information to the camera always overwrites the other pixels. As the
depth image is registered to the color image, the color pixels in
$\textbf{C}_a$ can also be mapped along with the depth pixels to
generate a virtual color image $\textbf{C}_b^*$ as well.

\begin{figure}
  \centering{\includegraphics[width=85mm]{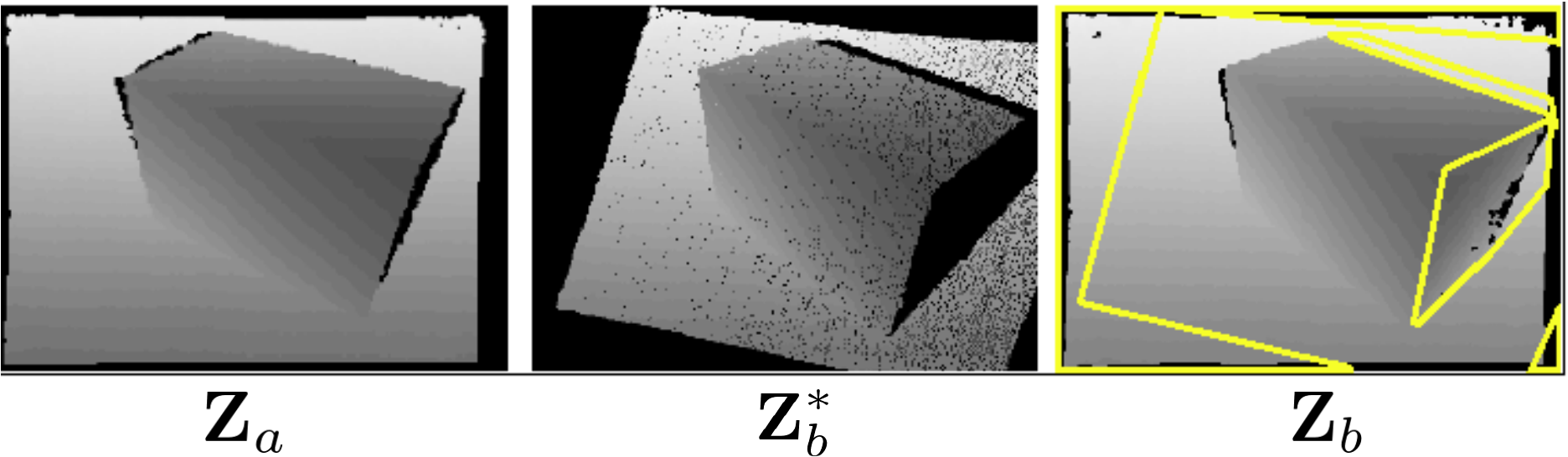}}
  \caption{An intuitive example of the prediction process. The depth
    image $\mathbf{Z}_{b}^{*}$ is synthetically generated from
    $\mathbf{Z}_{a}$ as the image captured by sensor $b$
    virtually. The uncorrelated information in $\mathbf{Z}_{b}$ is
    outlined with yellow lines.}
  \label{fig:forward}
\end{figure}
		
Then, the captured images $\textbf{Z}_{a}$, $\textbf{C}_{a}$, and
virtual images $\textbf{Z}_{b}^{*}$, $\textbf{C}_b^*$ are decomposed
into blocks of $8 \times 8$ pixels. In $\textbf{Z}_{b}^{*}$, some
blocks have no depth information due to the fact that none of the
pixels in $\textbf{Z}_a$ can be warped into these regions. This
indicates that the blocks with the same coordinates in $\textbf{Z}_b$
and $\textbf{C}_b$ contain the information which can only be observed
by sensor $b$. Sensor $a$ collects these block coordinates in the
set $\mathrm{B}_{p}$ and transmits to sensor $b$.

An illustration of this process is shown in Fig.~\ref{fig:forward}. In this
example, the regions in which the depth information can only be
observed by $\mathbf{Z}_{b}$ are outlined in yellow.

\subsubsection{Validation}
		
\begin{figure}
  \centering{\includegraphics[width=65mm]{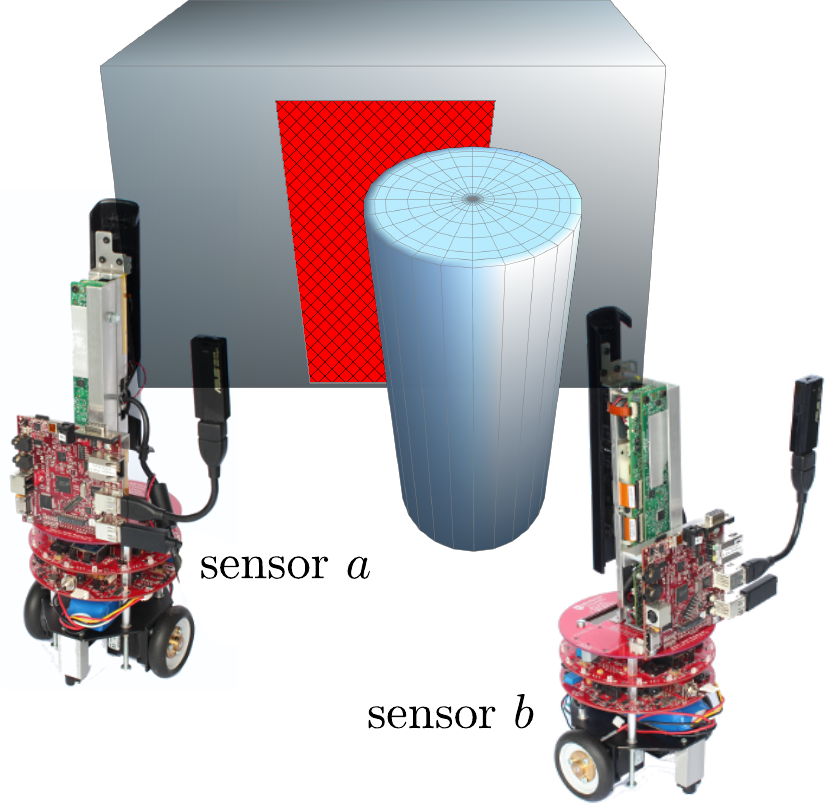}}
  \caption{The rectangular surface area at the background is within the FoV
    of sensor $b$, but occluded by the cylinder at the foreground.}
  \label{fig:occlusion}
\end{figure}
		
Although, in most circumstances, the prediction process can detect the
uncorrelated information in the images captured by the other sensor,
it may fail to operate correctly in situations when some points are
occluded by the objects that can be seen by sensor $b$, but not by
sensor $a$. A typical scenario is shown in Fig.~\ref{fig:occlusion}.
In this example, the cylinder is outside the FoV of sensor $a$, and
because of this, it falsely treats some parts of the background (the
dashed rectangular area) as the surface that is observable by sensor
$b$. However, since the surface of the cylinder is included in
$\textbf{Z}_{b}$, it occludes the background from the viewpoint of
sensor $b$. As a result, the prediction process cannot accurately
determine the uncorrelated depth and color information in this case.
		
In order to solve this problem, we include a validation mechanism into
the overall process. First, similar to the image warping process from
sensor $a$ to $b$, sensor $b$ generates the synthetic image
$\textbf{Z}_{a}^{*}$ as virtually captured from sensor $a$'s viewpoint
by mapping the pixels $(i_b,j_b)$ in $\mathbf{Z}_{b}$ to
$(i_{a},j_{a})$ in $\mathbf{Z}_{a}^{*}$ by applying
\begin{equation*}
  \begin{bmatrix}
    \frac{i_{a} - i_{c}}{f_{x}} & \frac{j_{a} - j_{c}}{f_{y}} & 1 &
    \frac{1}{z_{a}}
  \end{bmatrix}^{T} \!= \textbf{M}_{ab}^{-1}\!
  \begin{bmatrix}
    \frac{i_{b} - i_{c}}{f_{x}} & \frac{j_{b} - j_{c}}{f_{y}} & 1 &
    \frac{1}{z_{b}}
  \end{bmatrix}^{T}
\end{equation*}
In this process, the pixels representing the range information of the
surface of the cylinder move out of the image coordinate range and are
not shown in $\mathbf{Z}_{a}^{*}$. Sensor $b$ identifies the image
blocks containing these pixels, and records their coordinates in the
set $\mathrm{B}_{v}$. Then, sensor $b$ transmits only the image blocks
in $\textbf{Z}_b$ and $\textbf{C}_b$ that their coordinates are
included in the union of the sets $\mathrm{B}_{p}$ and
$\mathrm{B}_{v}$.

\subsection{Image Coding}
		
After the elimination of the redundant image blocks, the remaining
uncorrelated depth and color information is compressed to further
improve the communication channel usage.

For depth images, we use our own design \emph{Differential Huffman
  Coding with Multiple Lookup Tables (DHC-M)} lossless compression
scheme \cite{maxwang2016}. It is very fast and capable of compressing
the depth images without introducing any artifical refinements.

Among the many options for compressing color images, JPEG 2000
\cite{skodras2001} and H.264 \cite{wiegand2003} intra mode can be
mentioned as the leading schemes. As the wireless channels are
impacted by noise and being error prone, coding schemes that provide
progressive coding are considered to be more suitable for sensor
networks. Moreover, since a sensor node of a VSN has limited
computational capability, a lightweight image coding scheme is
required in sensor network applications. Progressive Graphics File
(PGF) scheme \cite{stamm2002}, which is based on discrete wavelet
transform with progressive coding features, has high coding efficiency
and low complexity. It has compression efficiency comparable to JPEG
2000, and is ten times faster. Moreover, PGF has a small, open source
and easy to use C++ codec \cite{libPGF} without any
dependencies. These properties make PGF suitable for onboard image
compression.

\subsection{Post-Processing at the Decoder Side}
\label{sec:filling}
		
\begin{figure}
  \centering{\includegraphics[width=90mm]{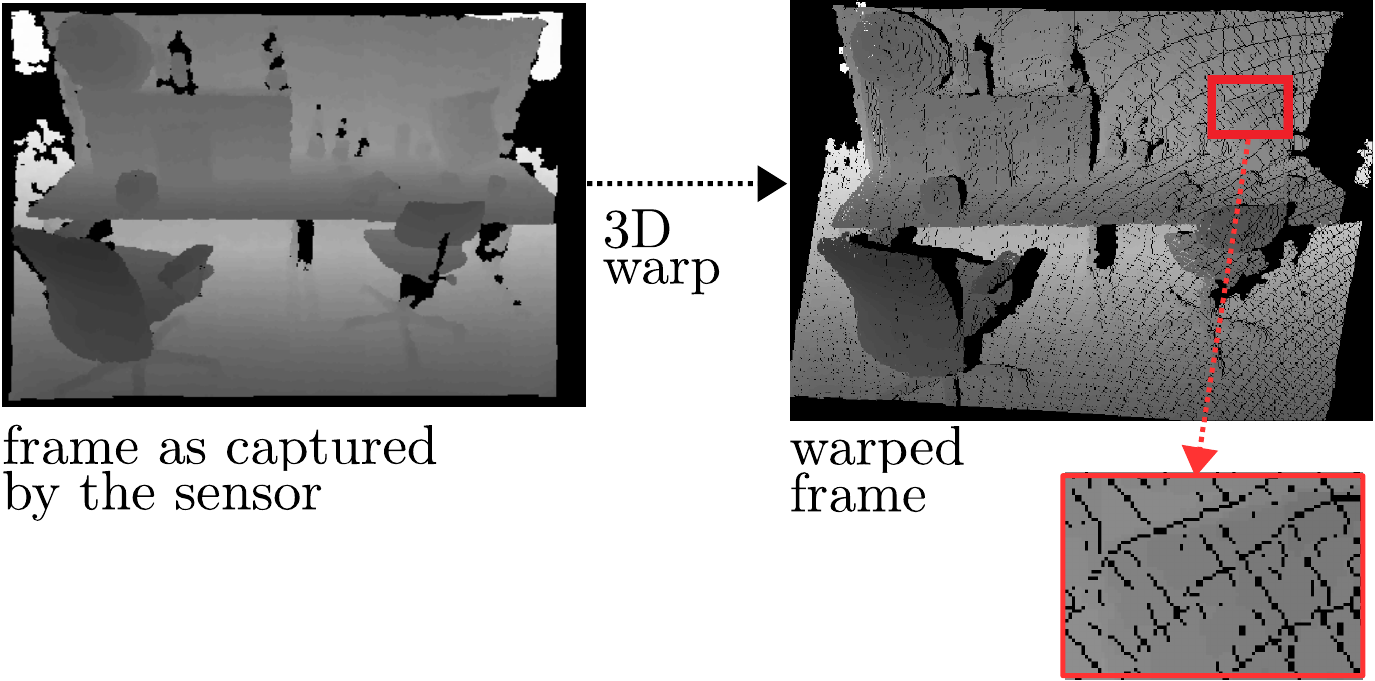}}
  \caption{Crack artifacts: Holes can be introduced during the image
    warping process due to the undersampling problem.}
  \label{fig:under_sample}
\end{figure}

\begin{figure}
  \centering{\includegraphics[width=70mm]{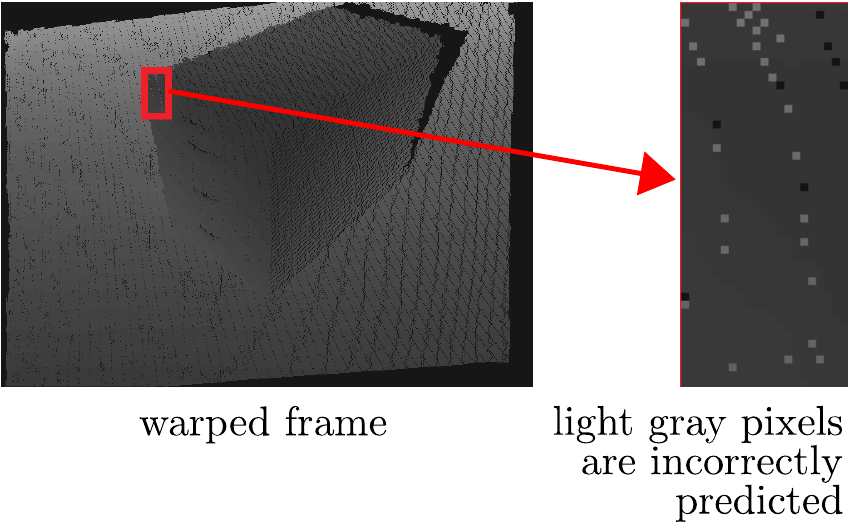}}
  \caption{Ghost artifacts: The light gray pixels actually belong to
    the background surface and falsely warped onto the surface at the
    foreground.}
  \label{fig:incorrect_prediction}
\end{figure}
		
At the decoder side (remote monitoring station), first, the received
bitstream is decompressed. Then, the color and
depth images captured by sensor $a$ are used to predict the color and
depth images captured by sensor $b$.
		
The 3-D image warping process (Eq. \ref{eq:warp}) may introduce some
visual artifacts in the synthesized view, such as
disocclusions\footnote{Disocclusions are areas occluded in the
  reference viewpoint and which become visible in the virtual
  viewpoint, due to the parallax effect.}, cracks\footnote{Cracks are
  small disocclusions, and mostly occur due to undersampling.}, or
ghosts\footnote{Ghosts are artifacts due to the projection of pixels
  that have background depth and mixed foreground/background
  color.}. Various methods have been proposed in the literature for
their prevention or removal \cite{Xi2013,Fickel2014}.
		
In our framework, as the information that can only be observed by
sensor $b$ is transmitted, disocclusions can be eliminated by filling
the areas affected by disocclusions in the synthesized image with the
color and depth information transmitted by sensor $b$. Then, the main
artifacts we need to deal with remain as cracks
(Fig.~\ref{fig:under_sample}) and ghosts
(Fig.~\ref{fig:incorrect_prediction}).

\subsubsection{Removal of Crack Artifacts}
\label{sec:remove-cracks}

The missing color information in cracks is frequently avoided by
operating a backward projection \cite{Mori2008}, which works in two
steps:
\begin{inparaenum}[(i)]
\item The cracks in the synthetic depth image are filled by a
median filter, and then a bilateral filter is applied to smoothen the
depth map while preserving the edges.
\item The filtered depth image
is warped back into the reference viewpoint to find the color of the
synthetic view.
\end{inparaenum}
This approach exhibits good performance on filling the cracks, but at
the same time it smoothens the complete image and introduces noise in
regions with correct depth values, especially on the object
boundaries. In order to avoid this adverse effect, we have modified it
by using an adaptive median filter. The filter is applied only on the
pixels with invalid depth values instead of the whole image. Instead
of warping back the complete image to find the color information, we
have adopted the work presented in \cite{Do2009} which warps back only the
filled pixels in cracks, because the color information of the other
pixels that are not in cracks can be directly estimated in the warping
process.

\subsubsection{Removal of Ghost Artifacts}

As illustrated in Fig.~\ref{fig:incorrect_prediction}, some background
surfaces are incorrectly shown on the foreground obstacle's
surface. This is because the pixels representing the foreground
surface become scattered after the warping process, and the background
surface can be seen through the interspaces between these pixels. In
order to remove this noise, we need to first identify the location of
the incorrectly predicted pixels and then fill them with the correct
values. As the value of the incorrectly predicted pixel is
significantly different from its neighboring pixels, this kind of
impulse noise can also be revised by using an adaptive median
filter. We propose a windowing scheme with a $3 \!\times\! 3$ pixels
size to determine whether or not a depth pixel contains incorrect
values. If more than half of the neighboring pixels are out of a
certain range, which is either much larger or much smaller than the
center pixel in the window, the center pixel is estimated as an
incorrectly predicted pixel. Then it is replaced with the median value
of its neighboring pixels which are not out of the range. The
corresponding color information can be found by backward warping,
which is similar to the solution for crack artifacts presented in
Section \ref{sec:remove-cracks}.

\section{Experimental Results and Performance Evaluation}
\label{sec:results}

In this section, we first evaluate the performance of the relative
pose estimation algorithm. Then, we analyze the overall performance of
the RPRR framework through the experiments conducted on our mobile VSN platform.

\subsection{Performance Evaluation of the Relative Pose Estimation}
\label{sec:relative}

In order to quantitatively evaluate the performance of the relative
pose estimation algorithm, we used two groups of datasets with varying
degrees of occlusions. We first generated our own datasets by using a
turntable setup to obtain the imagery viewed from accurately measured
angular positions.  A number of objects were placed on the center of
the turntable, and the images were captured with a tripod mounted
Kinect sensor. In the experiments with the first dataset group, the
ground truth is known exactly at every precisely controlled \ang{5}
interval. We used this setup to compare our algorithm (ICP-BD)
\cite{Wang2013} with the standard ICP \cite{Besl1992} and ICP in
inverse depth coordinates (ICP-IVD) \cite{Lui2012}. The performance of
the algorithms was evaluated based on the rotational and translational
RMS errors. The results show that
\begin{enumerate}[i.]
\item When the angular interval becomes greater than \ang{15},
  increasing amount of occlusion occurs between two sensors'
  views. Under such circumstances ICP-BD outperforms other variants as
  it reports much lower translational and rotational RMS error.
\item Standard ICP has the poorest performance across the
  experiments. ICP-IVD can provide similar accuracy in pose estimation
  before it diverges. However, as the scene becomes more occluded as
  the turntable is being rotated, ICP-IVD fails to converge sooner
  than ICP-BD.
\end{enumerate}
In summary, ICP-BD estimation accuracy is much better than that
of ICP and ICP-IVD. In addition, its estimation is very robust even
under large pose differences. Details of the experiment methodology
and results can be found in \cite{Wang2013}.

\begin{figure}
\centering{\includegraphics[width=85mm]{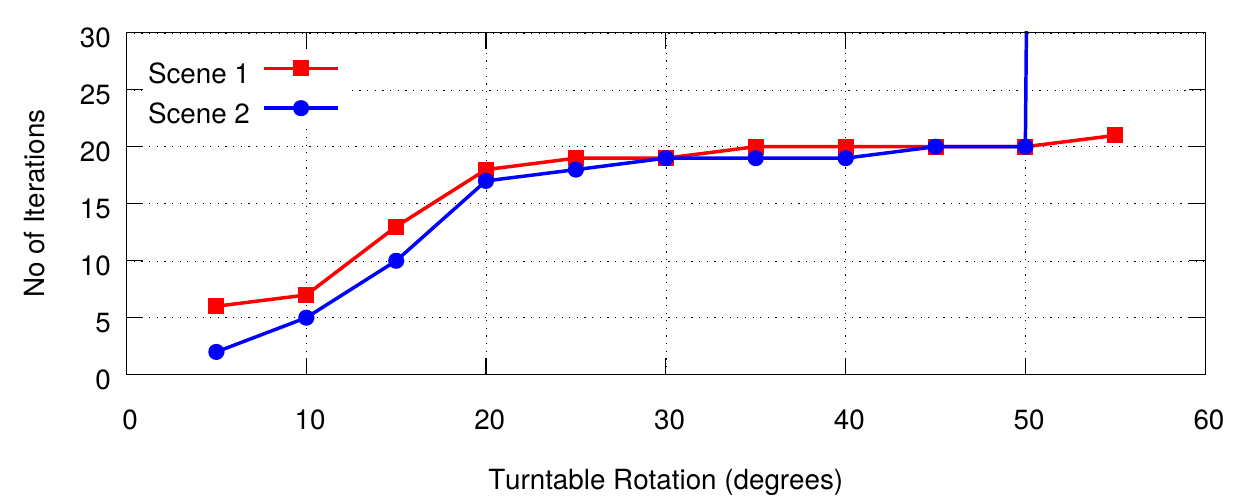}}
\caption{Number of iterations required by the ICP-BD algorithm in two
  scenes shown in Fig.~5 of \cite{Wang2013}.}
\label{fig:turn_iteration}
\end{figure}

\begin{figure}
  \centering{\includegraphics[width=75mm]{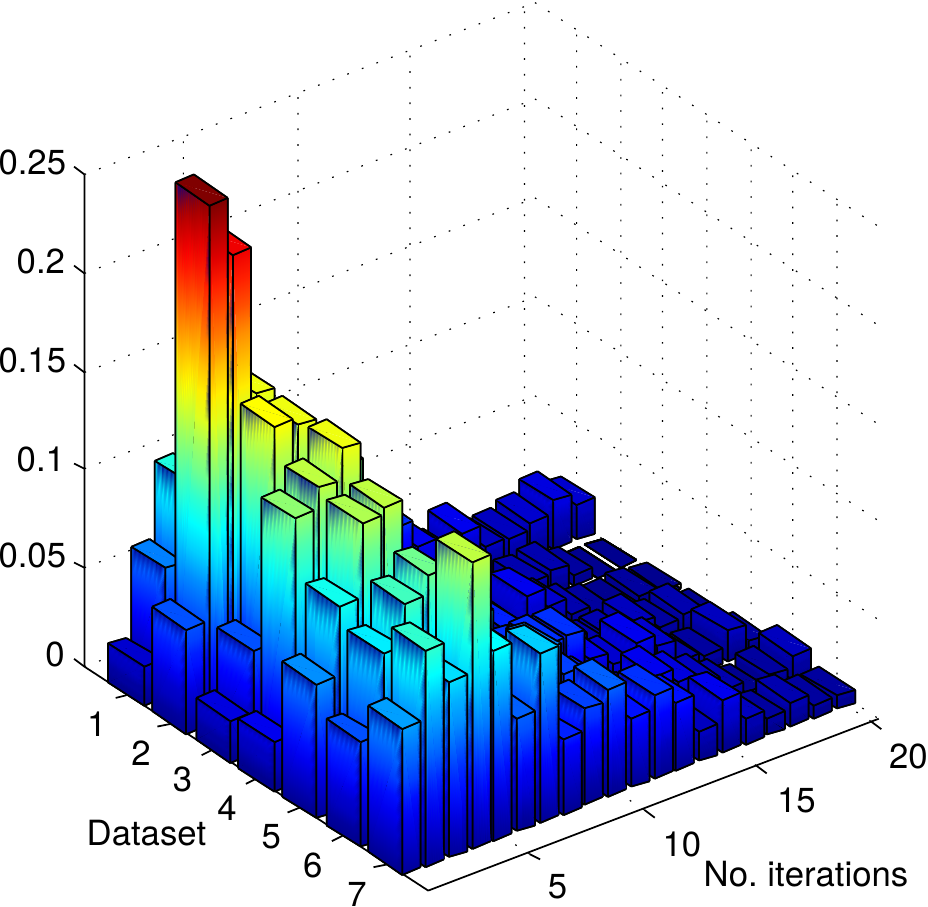}}
  \caption{Distributions of the number of iterations required
      for the convergence of the algorithm. Results were obtained over 
      the image pairs taken out of the sequences extracted from 
      the following seven RGB-D SLAM datasets:
       1.~{\small\textsf{freiburg1\_plant}},
       2.~{\small\textsf{freiburg2\_dishes}},
       3.~{\small\textsf{freiburg3\_cabinet}}, 
       4.~{\small\textsf{freiburg3\_large\_cabinet}},
       5.~{\small\textsf{freiburg3\_structure\_texture\_far}},
       \mbox{6.~{\small\textsf{freiburg3\_long\_office\_household}}, and}
       7.~{\small\textsf{freiburg1\_xyz\_cabinet}}~\cite{freiburg_dataset,Sturm12iros}.}
  \label{fig:histogram}
\end{figure}

We also evaluated the number of iterations required for the ICP-BD
algorithm's convergence. Our experiments show that, as one can expect,
the number of iterations increases as the angular difference between
two views increases. Two representative results are plotted in
Fig.~\ref{fig:turn_iteration}.

In order to gain further insight into the number of iterations
required by our algorithm in densely cluttered scenes, we used a
second group of datasets which were selected from the Technical
University of Munich Computer Vision Group's RGB-D~SLAM dataset and
benchmark collection~\cite{freiburg_dataset,Sturm12iros}. Each dataset
is a sequence of Kinect video frames capturing one scene from
different angles of view. For emulating the situations including
varying amounts of occlusion between two sensor views, we extracted
four new sequences from each dataset by taking one frame out of every
5, 10, 20, and 30 frames. For each trial we treated two consecutive
frames in the new sequence as the depth images captured by two
separate sensors. The number of iterations required by the ICP-BD
algorithm in trials that converged successfully are shown in
Fig.~\ref{fig:histogram}. The results show that the average number of
iterations is \num{5.1} and the maximum value is smaller than
20. Based on these numbers, we can say that the message exchange
complexity of the relative pose estimation algorithm is near-constant.
At each iteration, the depth information and image coordinates of 250
sampled points need to be transmitted, which lead to
\SI{1.09}{\kilo\byte} of bandwidth consumption approximately
(excluding the protocol overheads). Therefore, on average,
\SI{5.6}{\kilo\byte} of data are sent in each message when the
relative localization algorithm (Fig. \ref{fig:rel-pose-estimation})
distributed over sensors $a$ and $b$ is in operation.

\subsection{Performance Evaluation of the RPRR Framework}
\label{sec:bandwidth-energy}

\begin{figure*}[h!]
  \centering
  \subfloat[Images captured by sensor $a$.]{\includegraphics[width=0.8\linewidth]{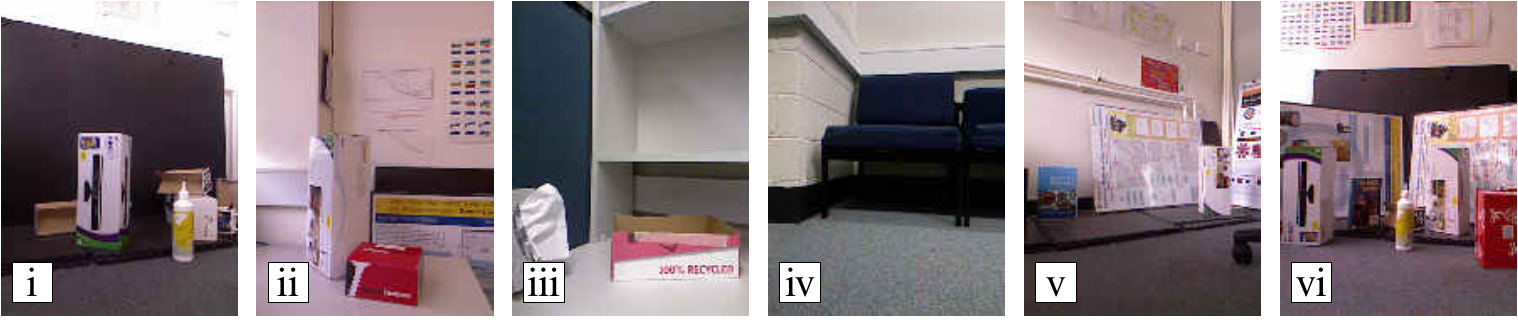}}\\
  \subfloat[Images captured by sensor $b$.]{\includegraphics[width=0.8\linewidth]{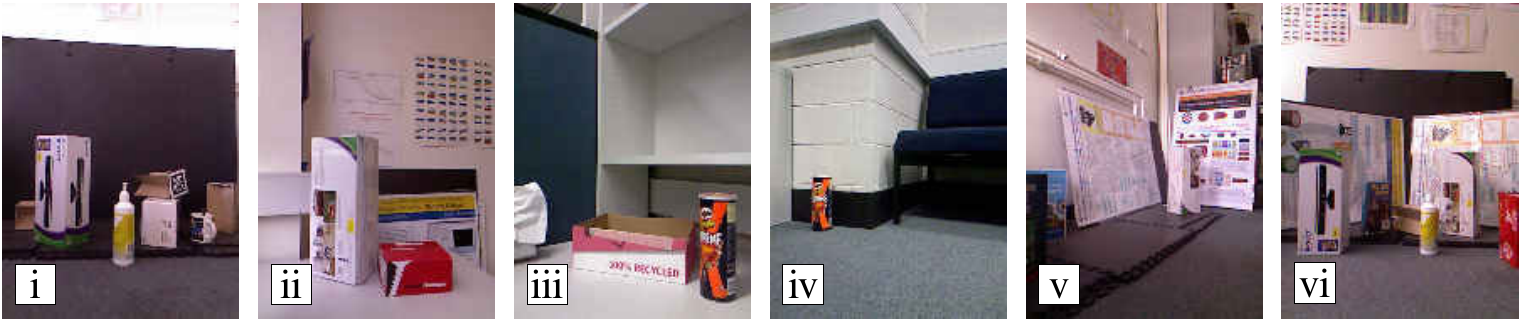}}\\
  \subfloat[Image blocks transmitted by sensor $b$ (black regions denote the image blocks that are not transmitted).]{\includegraphics[width=0.8\linewidth]{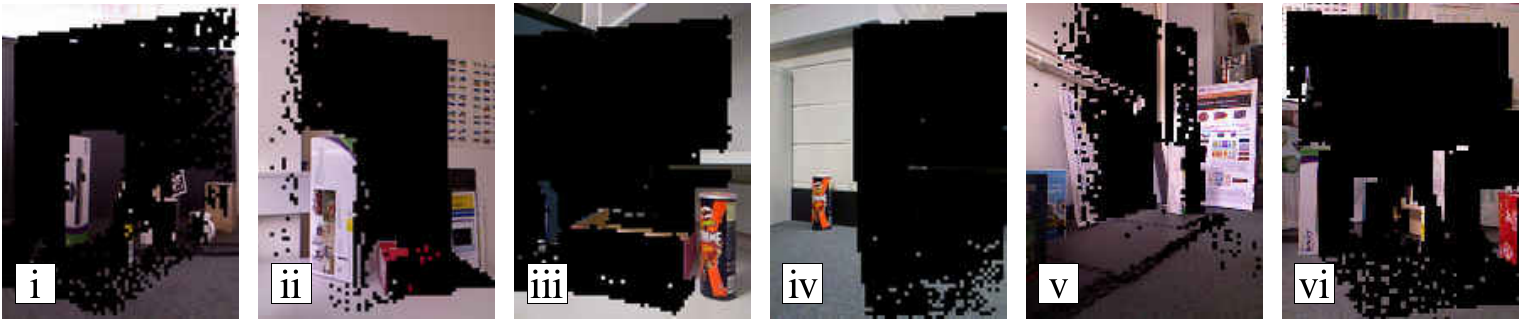}}\\
  \subfloat[Reconstructed images at the remote monitoring station.]{\includegraphics[width=0.8\linewidth]{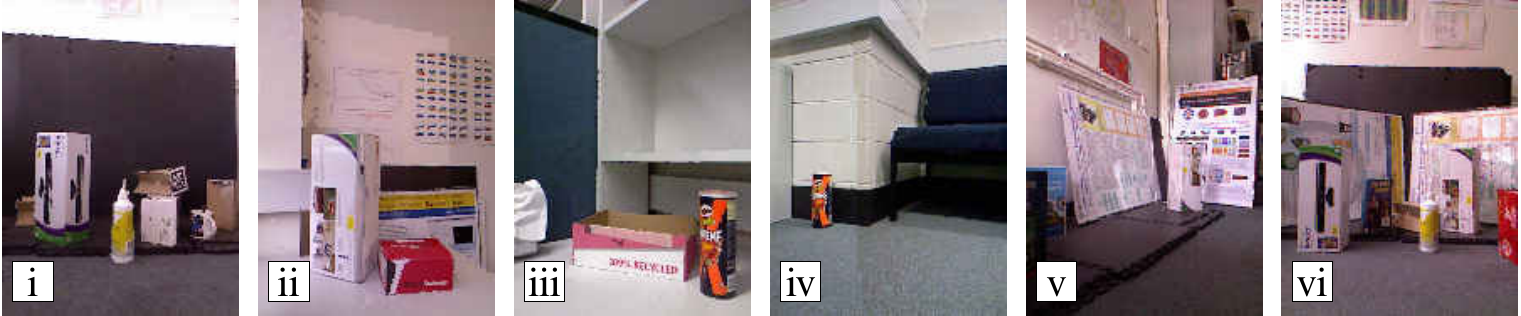}}\\
  \caption{A demonstration of the RPRR framework over six scenes.}
  \label{fig:sets}
\end{figure*}

\begin{figure}
  \centering\includegraphics[width=65mm]{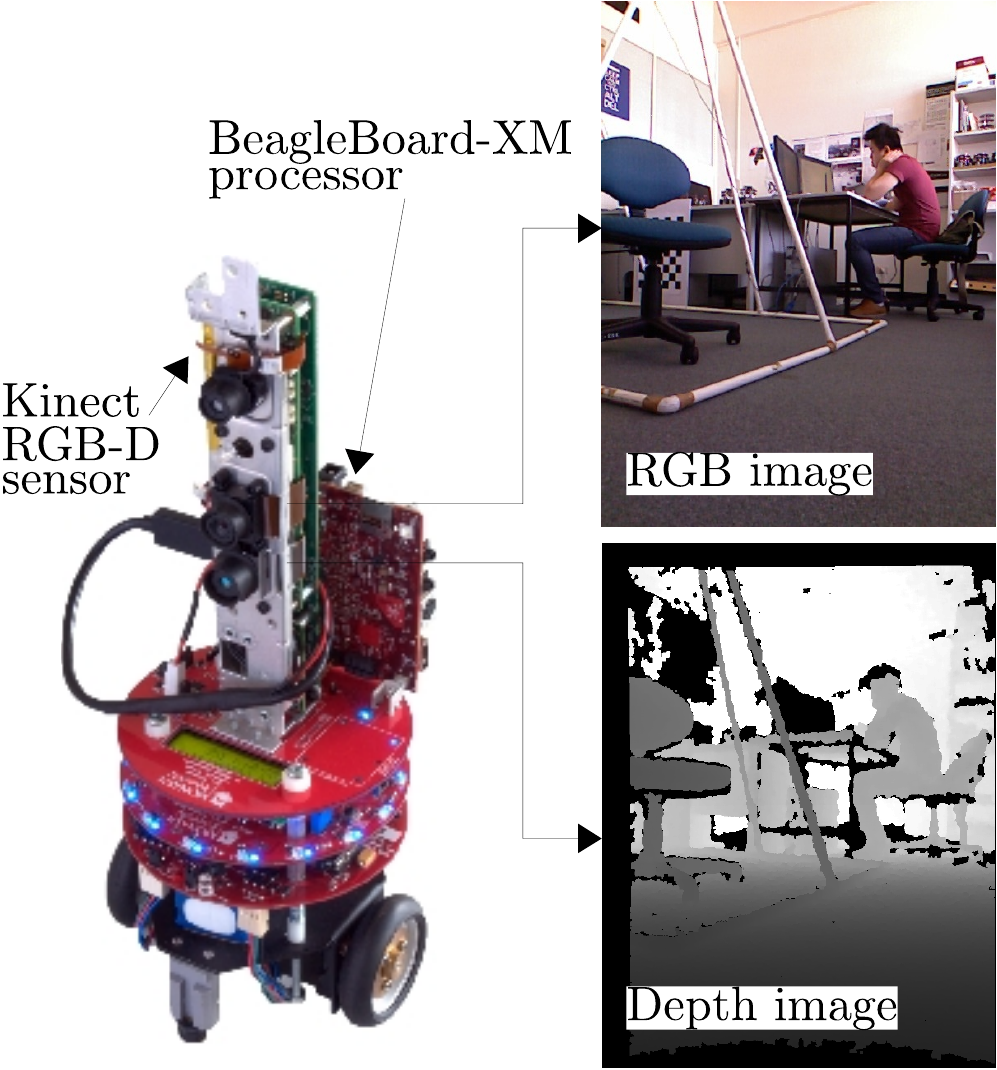}
  \caption{eyeBug \cite{dademo2011,eyebug02}, the mobile RGB-D sensor
    we use in our experiments. The color and depth data generated by
    the Kinect sensor is processed on a BeagleBoard-xM
    \cite{beagleboard} computer running the GNU/Linux operating
    system.}
  \label{fig:eyebug-kinect}
\end{figure}
		
In this set of experiments, we evaluated the performance of the RPRR
framework by using two mobile RGB-D sensors
(Fig. \ref{fig:eyebug-kinect}) of our VSN platform. The platform
consists of multiple mobile RGB-D sensors named ``eyeBug''
(Fig.~\ref{fig:eyebug-kinect}). EyeBugs were designed for computer
vision and mobile robotics experiments, such as multi-robot SLAM and
scene reconstruction. We selected the Microsoft Kinect as the RGB-D
sensor due to its low cost and wide availability. We mounted a Kinect
vertically at the center of the top board of each eyeBug. A Kinect is
capable of producing color and disparity-based depth images at a rate
of \num{30} frames/second. A BeagleBoard-xM single-board computer
\cite{beagleboard} was used for image processing tasks. Each
BeagleBoard-xM has a 1 GHz ARM Cortex-A8 processor, a USB hub, and a
HDMI video output port. A USB WiFi adapter was connected to the
BeagleBoard to provide communication between robots. We ran an
ARM-processor-optimized Linux kernel. OpenKinect \cite{openkinect},
OpenCV \cite{opencv} and libCVD \cite{libcvd} libraries were installed
to capture and process image information. The default RGB video stream
provided by the Kinect uses 8 bits for each color at VGA resolution
(640 $\times$ 480 pixels, 24 bits/pixel). The monochrome depth video
stream is also in VGA resolution. The value of each depth pixel
represents the distance information in millimeters. Invalid depth
pixel values are recorded as zero, indicating that the RGB-D sensor is
not able to estimate the depth information of that point in the 3-D
world.
		
Color and depth images were captured in six different scenes, as shown
in Fig.~\ref{fig:sets}. In this set-up, sensor $a$ transmits entire
captured color and depth images to the central monitoring
station. Then, sensor $b$ is required to transmit only the
uncorrelated color and depth information that cannot be observed by
sensor $a$. At the central monitoring station, the color and depth
images captured by sensor $b$ are reconstructed by using the
information transmitted by two sensors.
As the color and depth images captured by sensor $a$ are
compressed and transmitted to the receiver in their entirety, we only needed to evaluate
the reconstruction quality of the images captured by sensor $b$. The
depth images are usually complementary to the color images in many
applications, and in our framework the color images are reconstructed
according to depth image warping. So, if the color
images can be accurately reconstructed, so the reconstructed depth
images as well. Therefore, in this set of experiments we focused on
evaluating the quality of the reconstructed color images.

\subsubsection{Subjective Evaluation}
		
The image blocks transmitted by sensor $b$ are shown in the third row
of Fig.~\ref{fig:sets}. In the fourth row of the figure reconstructed
images can be seen. They were obtained by stitching the blocks
extracted from the warped sensor $a$ images into the black regions of
the corresponding sensor $b$ images. In the reconstructed images of
scenes 2 and 4, we observe significant color changes on the stitching
boundary. This is caused by the illumination variations within the
scene, and auto-iris response of the sensors to different levels of
scene brightness.

Generally, it is clear that the reconstructed images preserve the
structural information of the original images accurately.

\subsubsection{Objective Evaluation}

\begin{figure*}
  \centering
  \subfloat[Scene 1]{\includegraphics[width=60mm]{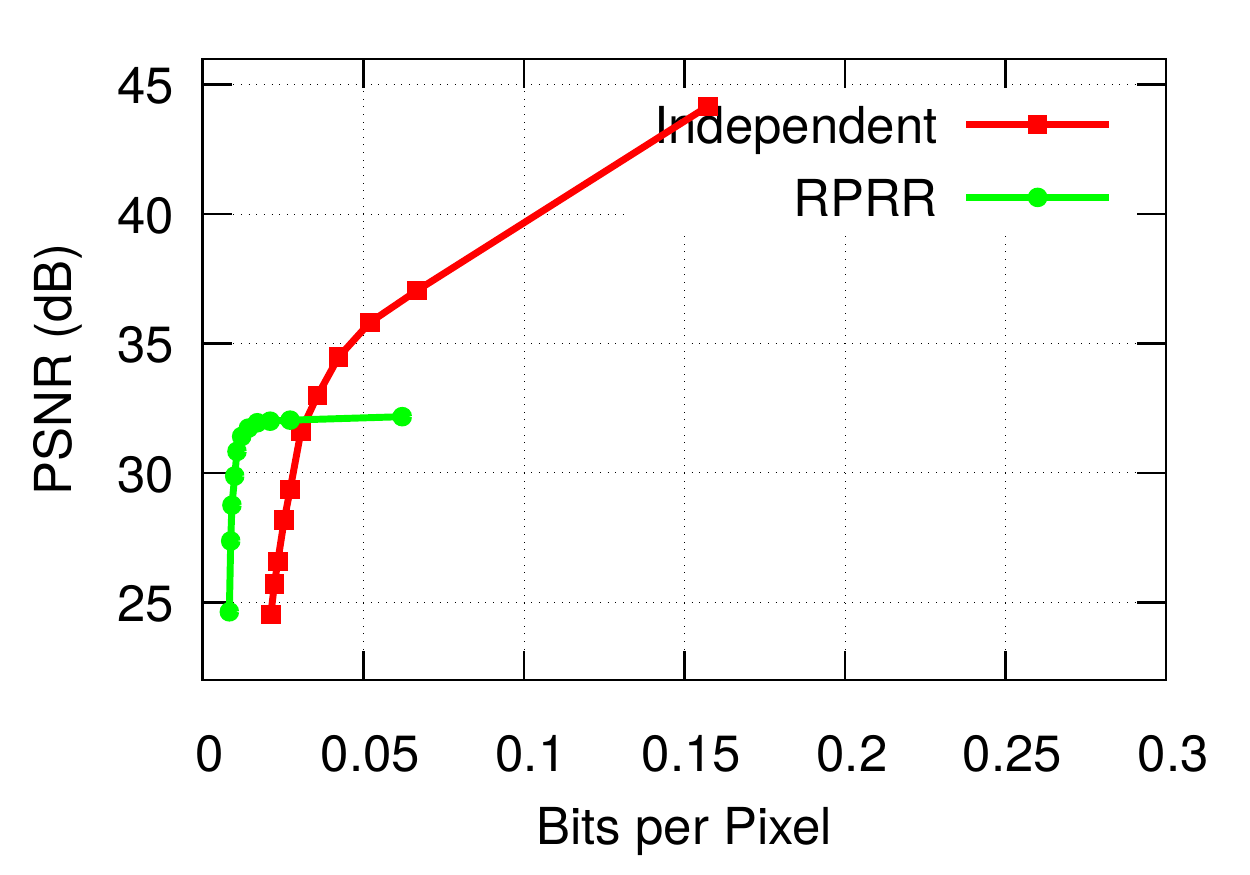}}
  \subfloat[Scene 2]{\includegraphics[width=60mm]{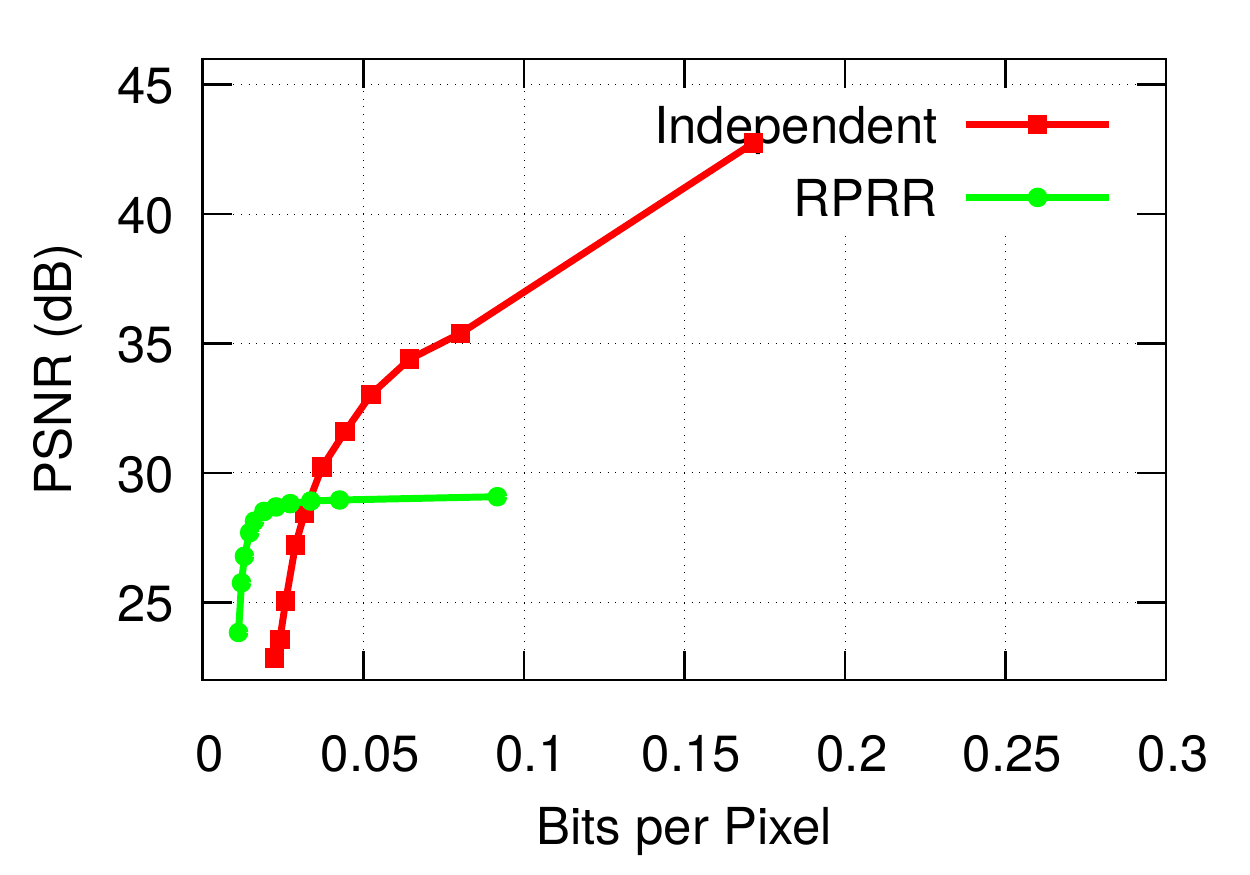}}
  \subfloat[Scene 3]{\includegraphics[width=60mm]{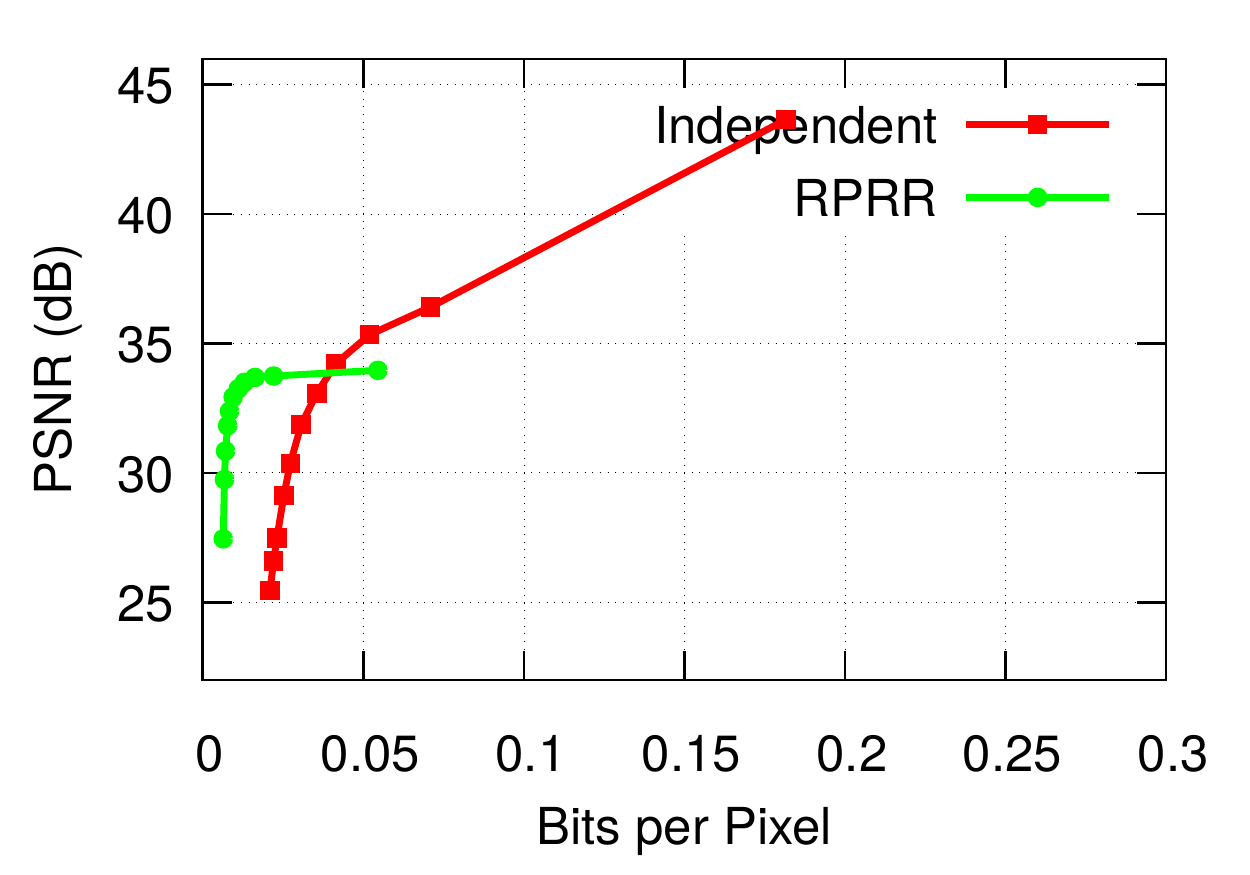}}\\%
  \subfloat[Scene 4]{\includegraphics[width=60mm]{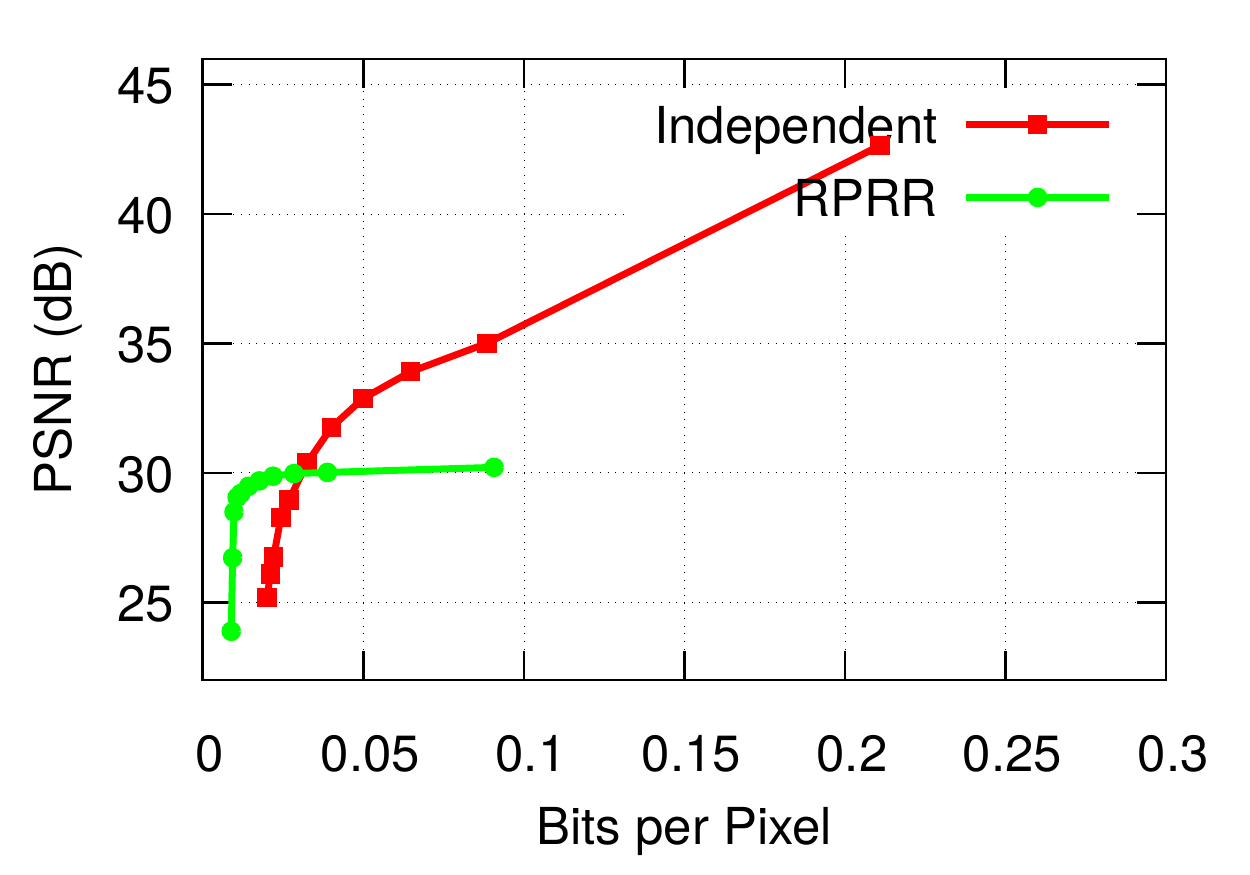}}
  \subfloat[Scene 5]{\includegraphics[width=60mm]{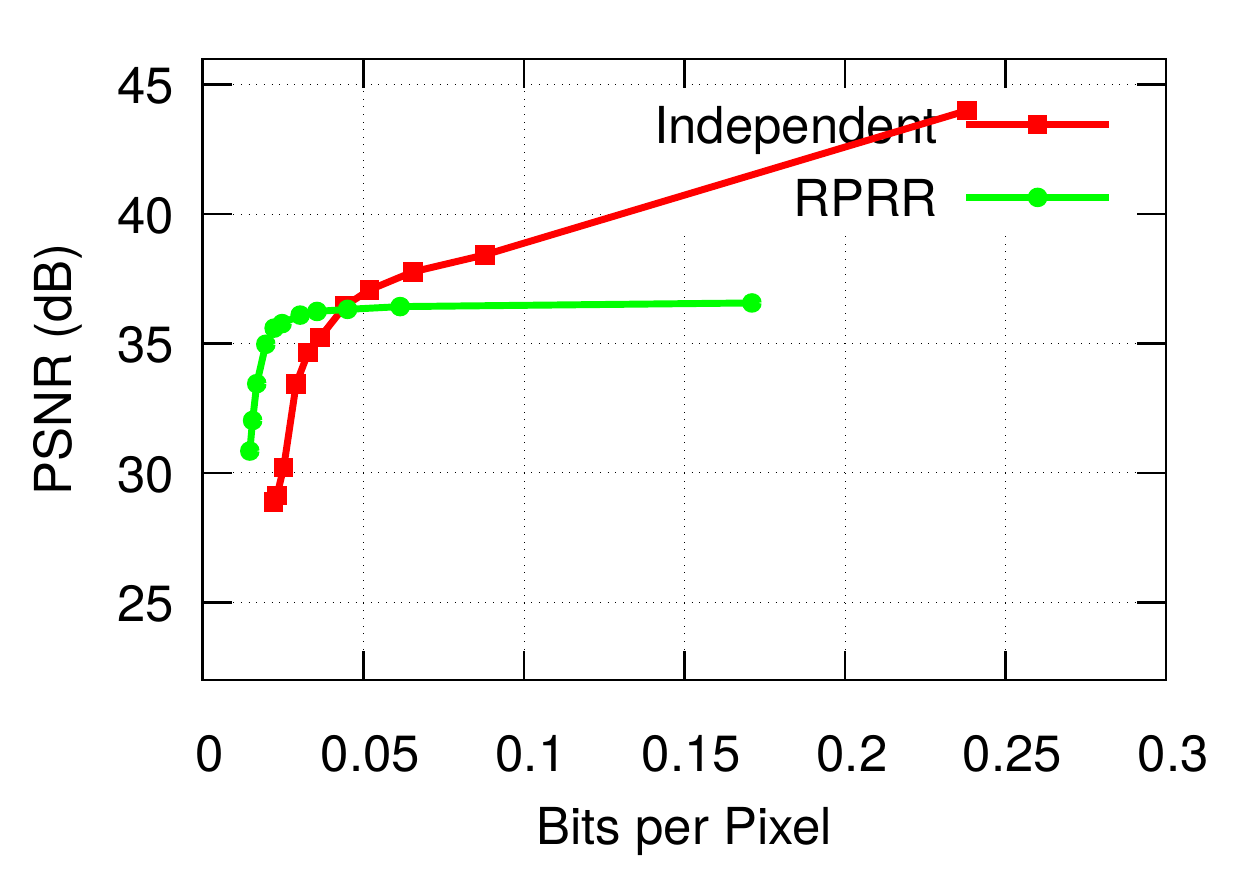}}
  \subfloat[Scene 6]{\includegraphics[width=60mm]{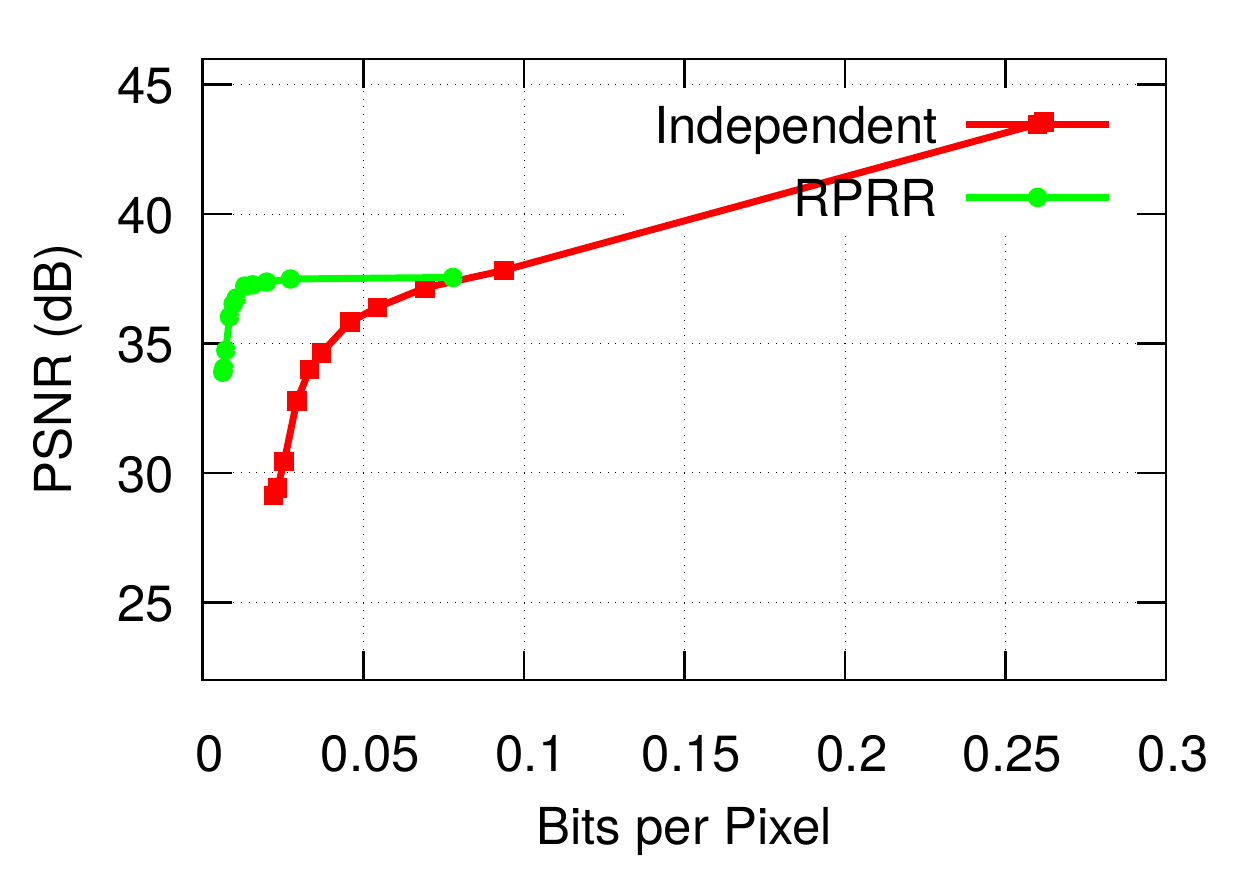}}
  \caption{Comparisons of PSNR (dB) achieved by compressing the images
    at various levels by using the RPRR framework
    against transmitting them independently.}
  \label{fig:curves}
\end{figure*}

Even though many approaches have been proposed to compress multi-view images
\cite{Lu2007,Chia2012,Colonnese2012,Wu2007,Deli2012,Gehrig2007,Merkle2009},
they cannot be applied in our system. These approaches either
require the transmitter to have the knowledge of the full set of
images or only work on cameras with very small motion
differences. In contrast, in our case, each sensor only has its own captured
image, and the motion difference between two visual sensors is very
large. 
To the best of our knowledge, our proposal is the first distributed
framework that efficiently codes and transmits images captured by multiple
RGB-D sensors with large pose differences, and so, we do not have any work to
compare ours against. For this reason, we can only compare the
performance of our framework with the approaches which compress and
transmit images independently.

As the color information is coded using PGF \cite{stamm2002} lossy mode, we can vary 
the compression ratio, and consequently, coding performance. The
performance was evaluated according to two aspects: reconstruction
quality and Bits per pixel (bpp). We measured the
Peak-Signal-to-Noise-Ratio (PSNR) between the reconstructed and
original images captured by sensor $b$ with different bpp. The results are shown in Fig.~\ref{fig:curves}. %

Fig.~\ref{fig:curves} shows that the RPRR framework can achieve much
lower bpp than the independent transmission scheme. However, the PSNR
upper bounds achieved by RPRR framework are limited. It is because the
reconstruction quality depends on the depth image accuracy and
correlations between the color images.  Since the depth images
generated by a Kinect sensor are not accurate enough, the displacement
distortion of depth images, especially the misalignment around the
object edges, introduce noise in the reconstruction process. Another
reason is the inconsistent illumination between the color images
captured by two sensors. Even if the prediction and validation
processes establish the correct correspondences between two color
pixels according to the transformation between depth images, the
values of these two color pixels can be very different due to the
various brightness levels in two images. These characteristics lead to
low PSNR upper bounds of the reconstructed color images.  Several
methods \cite{Vijayanagar2012,Vijayanagar2014} have been proposed to
overcome this drawback, however the time-complexity of these methods
prevents them from being implemented on sensor systems with
constrained computational resources. We can see that the reconstructed
color image in Scene 6 has the highest PSNR, it is because the
relative pose between two sensors is small, which leads to small
differences in the structure of the captured scenes and the brightness
of their captured images. Therefore, more information captured by
sensor $b$ can be reconstructed by information observed by sensor
$a$. For that reason, according to Fig.~\ref{fig:sets}\,(a-vi), only a
small number of blocks in image captured by sensor $b$ need to be
transmitted. We also observe that Scenes 2 and 4 have the lowest
reconstruction qualities, this is because the brightness level is
quite different in the color images captured by two sensors (see image
pairs shown in Figs.~\ref{fig:sets}\,(a-ii)(b-ii), and
Figs.~\ref{fig:sets}\,(a-iv)(b-iv)). Although the structures of the
scenes are preserved nicely in the reconstructed color images,
distinct color changes over the stitching boundaries are shown in
Figs.~\ref{fig:sets}\,(d-ii) and
~\ref{fig:sets}\,(d-iv). Consequently, we can say that the RPRR
framework is suitable for implementation of the VSN applications with
very limited bandwidth requiring very high compression ratios.  This
is because when the bpp or the compression ratio increases, the
quality of the color image reconstructed by RPRR decreases more
gradually than the quality of the image compressed by the independent
transmission scheme. %

\subsubsection{Energy Consumption} 
		
The limited battery capacity of mobile sensors places limits on
their performance. Therefore,  a data transmission scheme, while attempting to
reduce the transmission load, must not have a significant negative
impact on the overall energy consumption.  In this section, we present
our experimental measurements and evaluation regarding the overall
energy consumption and amount of transmitted data of the RPRR
framework collected on our eyeBug mobile visual sensors to demonstrate
this aspect.
		
The overall energy consumption of the RPRR framework can be measured
by
\begin{align}
  E^R_{\text{overall}}& = E_{\text{processing}}+E_{\text{encoding}}+E_{\text{sending}}\nonumber\\
  & = V_{o}I_{p}t_p+V_{o}I_{e}t_e + V_{o}I_{s}t_{s}
\end{align}
in which $V_{o}$ denotes the sensor's operating voltage, and $I_p,
I_e,$ and $I_s$ represent the current drawn from the battery during
processing, encoding, and sending operations. $t_p, t_e,$ and
$t_s$ are the corresponding operation times required for these
procedures.

The overall energy consumption when images are transmitted
independently can be measured as,
\begin{align}
  E^I_{\text{overall}}& = E_{\text{encoding}}+E_{\text{sending}}\nonumber\\
  & = V_{o}I_{e}t_e + V_{o}I_{s}t_{s}.
\end{align}
Note that, the operation times $t_e$ and $t_s$ are different in the two
transmission schemes as the image sizes change after removing the
redundant information.

Our sensor operates at \SI{15}{\volt}, and the current levels remain
fairly constant during each operation. We measured them as follows:
$I_p = \SI{0.06}{\ampere}$, $I_e = \SI{0.06}{\ampere}$, and
$I_s = \SI{0.12}{\ampere}$. Our experiments show that in the RPRR
framework, due to different compression ratios, the transmission time
varies between \SI{32} and \SI{42}{\milli\second}, and the operational
time for processing and encoding remains between \SI{509} and
\SI{553}{\milli\second}. The overall energy consumption of the RPRR
scheme changes between \SI{480} and \SI{520}{\milli\joule}, depending
on the compression ratio. The corresponding values for the independent
scheme are between \SI{918} and \SI{920}{\milli\joule}. The data
clearly show that the RPRR framework leads to the consumption of much
lower battery capacity than the independent transmission scheme. It
cuts the overall energy consumption of the sensor nearly by half. In
the RPRR framework, the energy consumption for two sensors are
asymmetric, and if sensor $a$ always transmits complete images, its
energy will be quickly drained. A simple method to prolong the network
lifetime is for the two sensors to transmit complete images
alternately. The current consumed by an eyeBug in idle state is
\SI{650}{\milli\ampere}. According to the experimental results above,
the theoretical operational time of RPRR on a pair of eyeBugs with
\SI{2500}{\milli\ampere\hour} 3-cell (\SI{11.1}{\volt}) LiPo batteries
is around \SI{5.2} hours. In this period, around \num{3.24e4} color
and depth image pairs can be transmitted to the remote monitoring
station.

\subsubsection{Transmitted Data Volume} 

\begin{figure*}%
  \centering
  \subfloat[Scene 1]{\includegraphics[width=60mm]{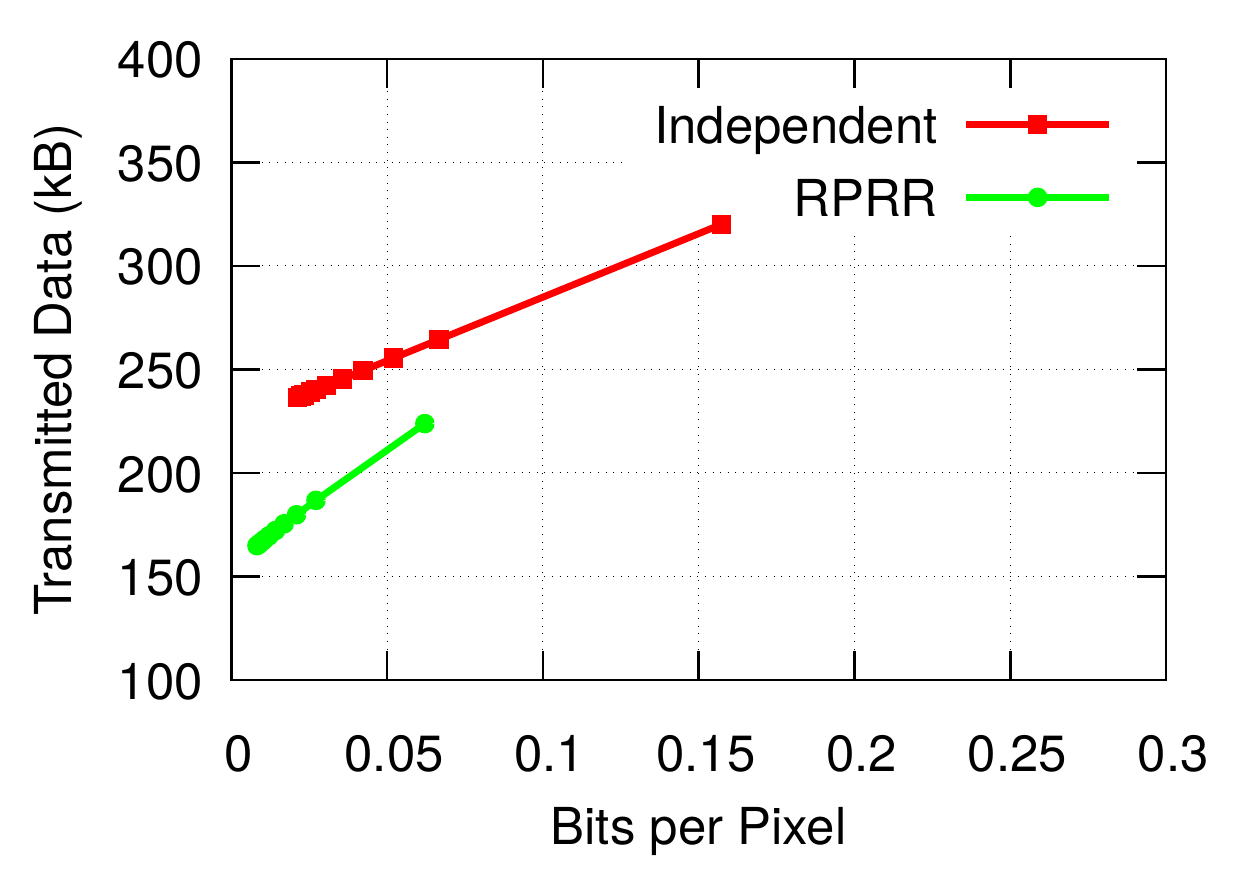}}
  \subfloat[Scene 2]{\includegraphics[width=60mm]{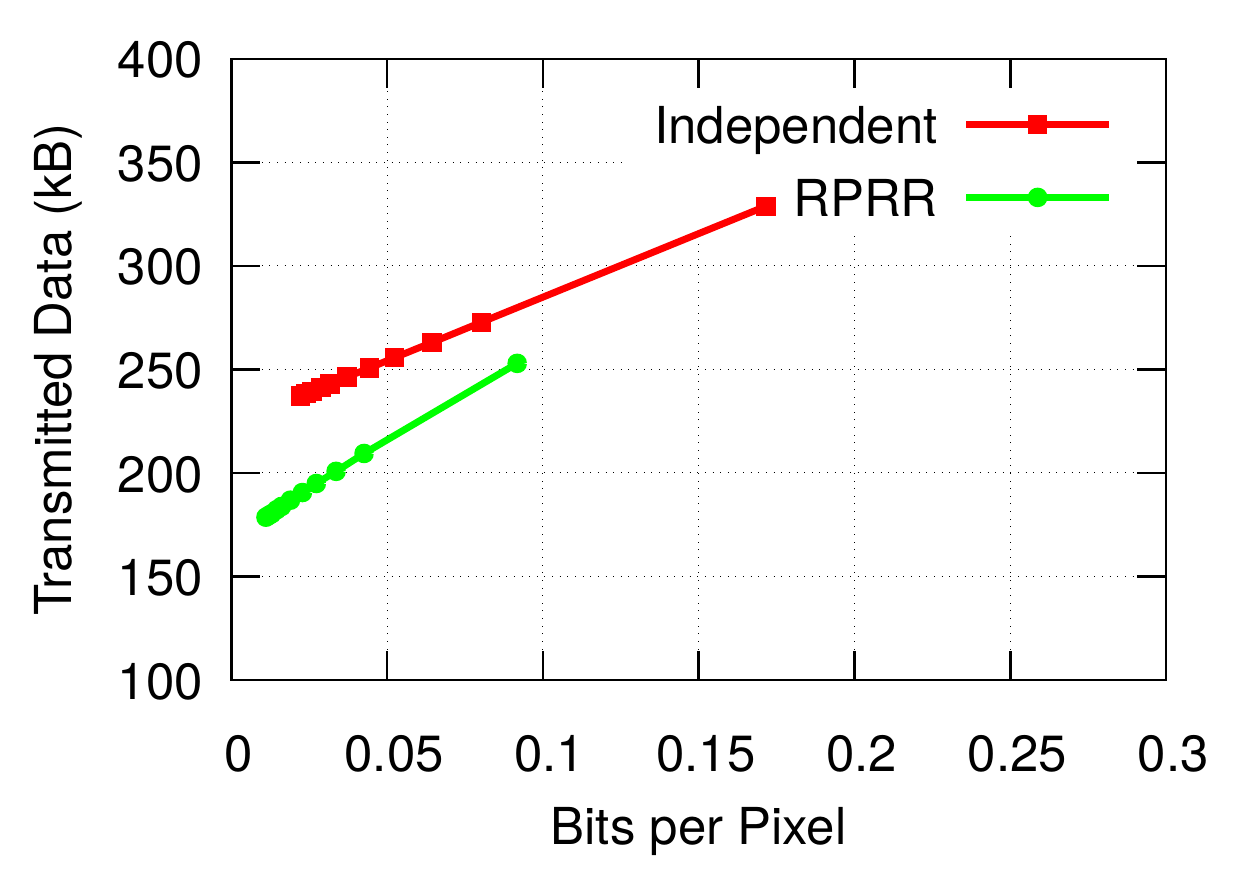}}
  \subfloat[Scene 3]{\includegraphics[width=60mm]{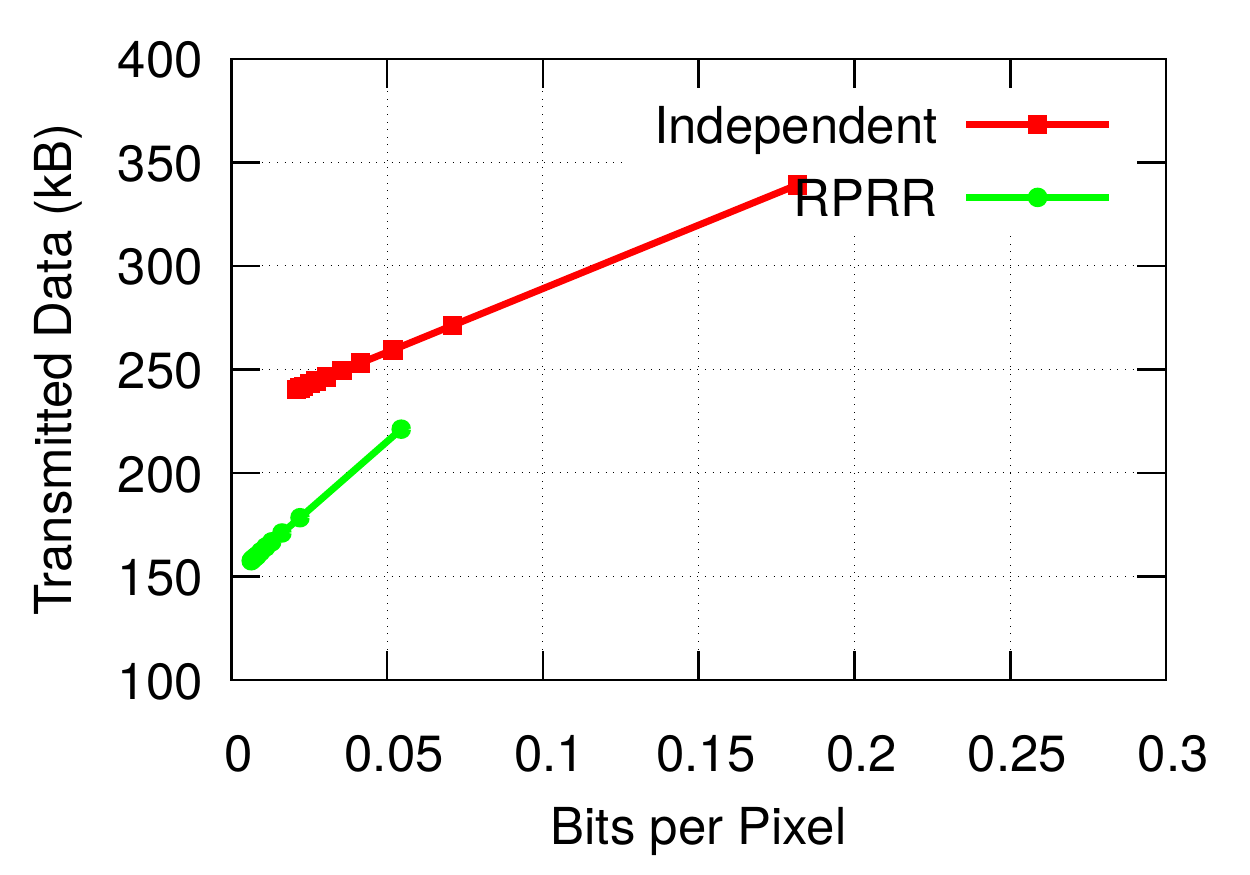}}\\%
  \subfloat[Scene 4]{\includegraphics[width=60mm]{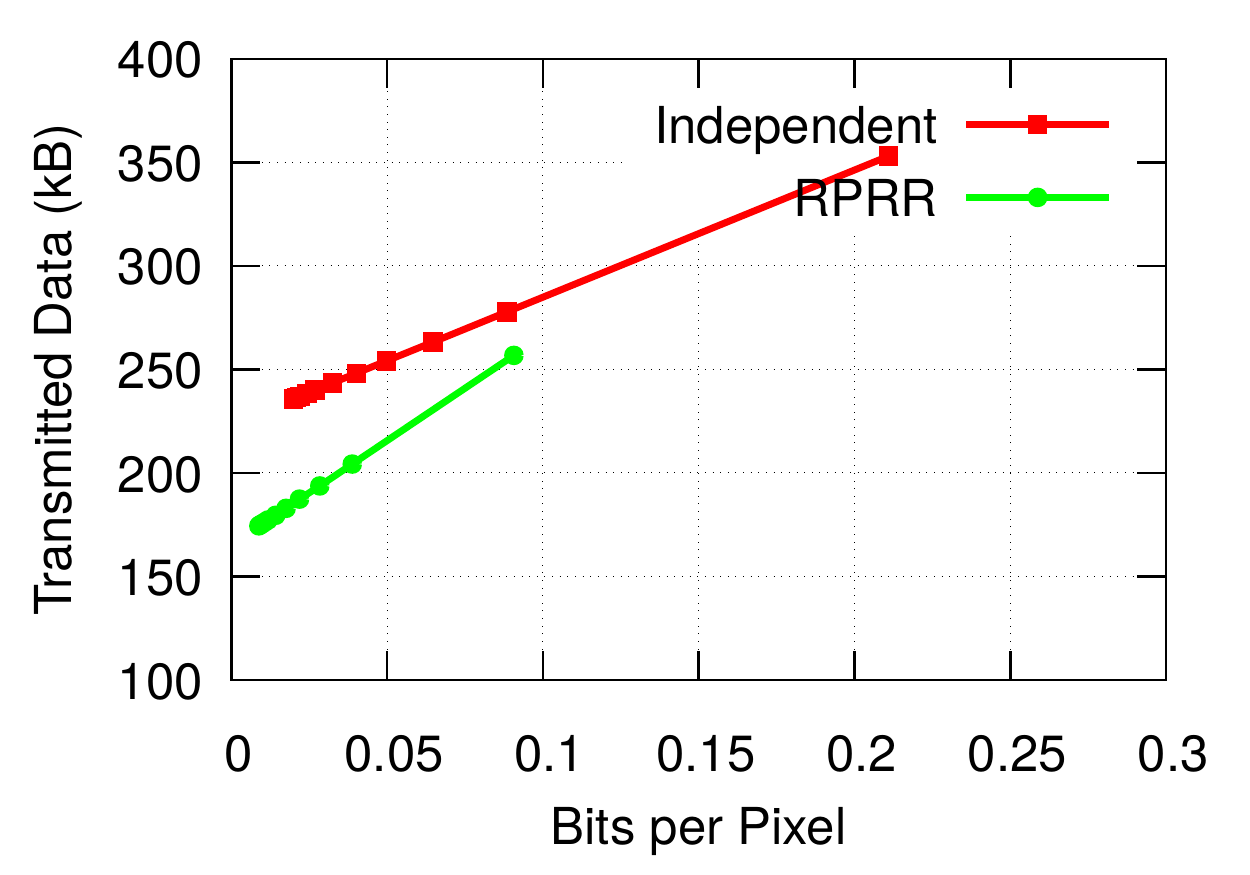}}
  \subfloat[Scene 5]{\includegraphics[width=60mm]{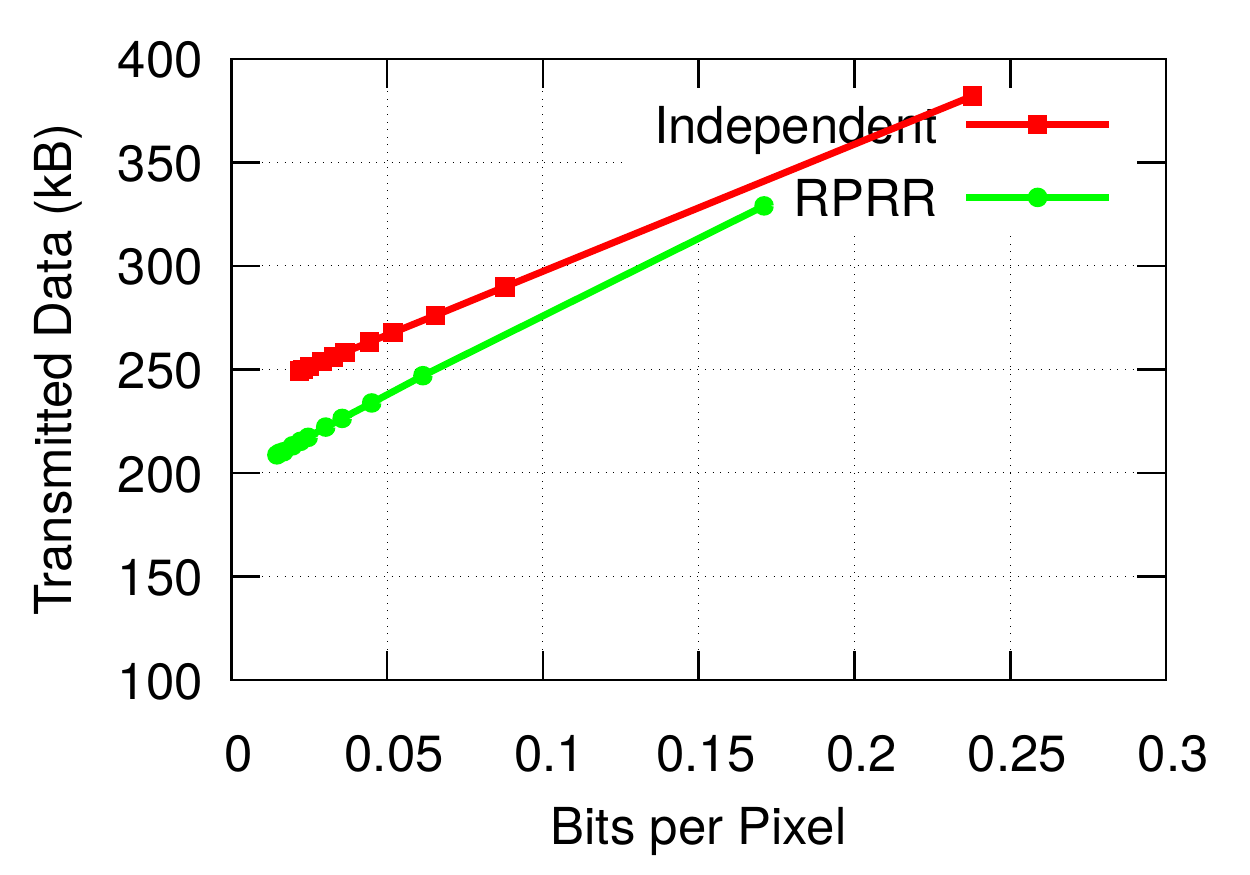}}
  \subfloat[Scene 6]{\includegraphics[width=60mm]{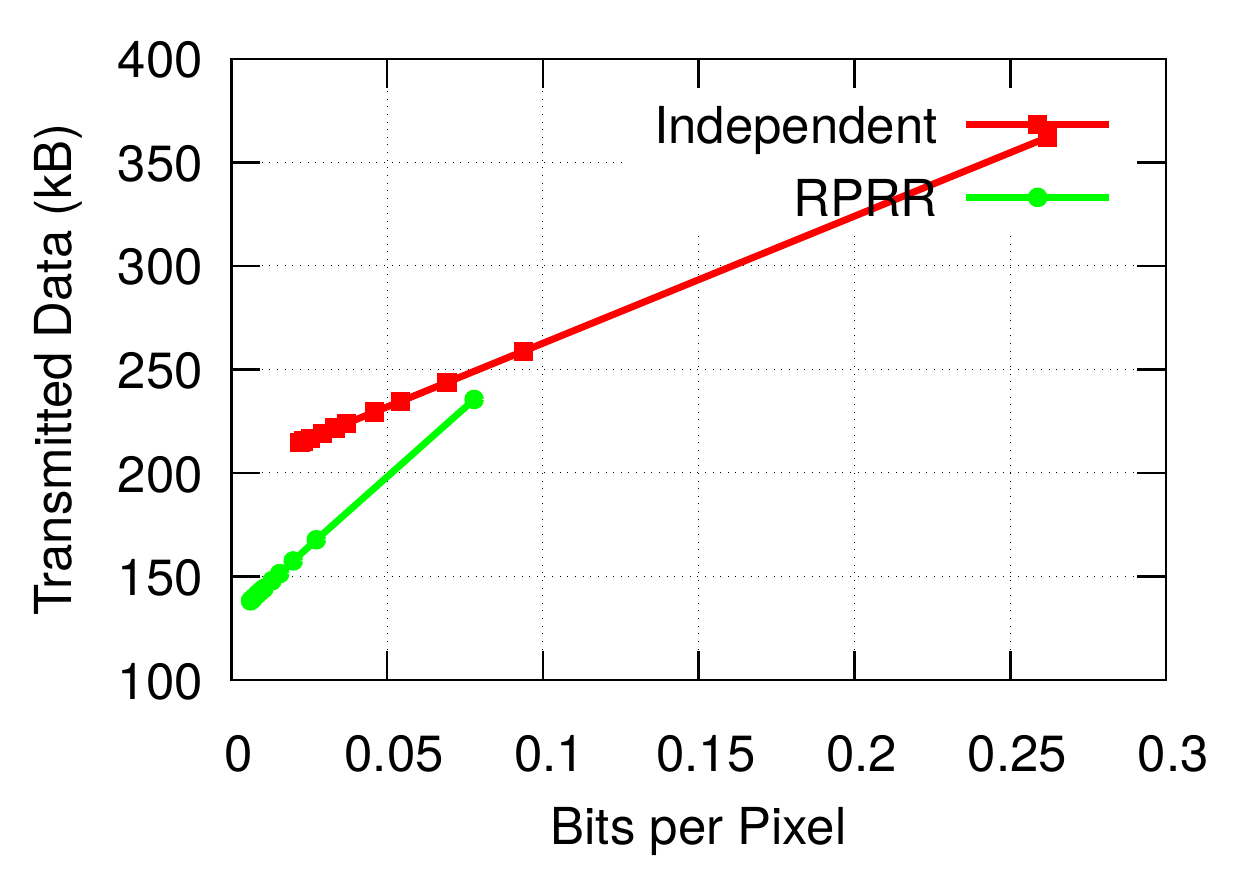}}
  \caption{Comparisons of the transmitted data for color images at
    various compression levels by using the RPRR framework against
    transmitting them independently.}
  \label{fig:curves-bw}
\end{figure*}

Finally, we compare the amount of transmitted data for two pairs of
color and depth images required by the RPRR and the independent
transmission schemes. The results are shown in
Fig.~\ref{fig:curves-bw}. We can see that, bits per pixel achieved by
the independent transmission approach is much higher than bit per
pixel achieved by the RPRR framework.  It is also noticeable that even
if the bits per pixel required by a color image is the same in both
approaches, the RPRR framework transmits fewer number of bytes. This
is because only parts of the color and depth images need to be
transmitted in RPRR. In contrast, a complete depth image has to be
sent in the independent transmission scheme. The data clearly show
that the RPRR framework leads to more efficient use of the wireless
channel capacity than the independent transmission scheme.

\section{Concluding Remarks and Discussion}

We presented a novel collaborative transmission framework for mobile
VSNs that efficiently removes the redundant visual information
captured by RGB-D sensors. The scheme, called \emph{Relative Posed
  based Redundancy Removal (RPRR)}, considers a multiview scenario in
which pairs of sensors observe the same scene from different
viewpoints. Taking advantage of the unique characteristics of depth
images, our framework explores the correlation between the images
captured by these sensors using solely the relative pose
information. Then, only the uncorrelated information is
transmitted. This significantly reduces the amount of information
transmitted compared with sending two individual images
independently. The scheme's computational resource requirements are
quite modest, and it can run on battery-operated sensor
nodes. Experimental results show that the compression ratio achieved
by the RPRR framework is $2 \frac{1}{2}$ times better than the
independent transmission scheme, and it yields this result while
nearly halving the energy consumption of the independent transmission
scheme on average.

The RPRR framework is the first attempt to remove the redundancy in
the color and depth information observed by VSNs equipped with RGB-D
sensors, and so there is room for further improvements. For example,
our scheme only operates on pairs of mobile sensors at this stage. A
simple extension of the RPRR framework for networks with a large
number of RGB-D sensors is to choose one sensor as the reference which
transmits complete images (like sensor $a$ in Fig. \ref{fig:overview})
while the other sensors transmit only the uncorrelated information
(like sensor $b$ in Fig. \ref{fig:overview}).  However, a certain
amount of redundancy still exists in this approach and further
refinements are possible.
Our future research efforts will concentrate on developing a more
sophisticated extension which uses feature matching algorithms to
assign sensors with overlapping FoVs to the same subgroups and applies
RPRR on sensors in the same subgroup to remove redundancies in
networks with a large number of RGB-D
sensors.  %

\bibliographystyle{IEEEtran} 
\bibliography{redundancy-removal}
\vspace*{-15mm}
\begin{IEEEbiographynophoto}{Xiaoqin
    Wang} graduated with a B.Sc. (Hons.) First Class degree in
  Electronics and Communications Engineering from the University of
  Birmingham, UK. He then obtained his Communication and Signal
  Processing M.Sc. degree from Imperial College London, UK, in
  2011. He is currently pursuing his Ph.D. degree at the Department of
  Electrical and Computer Systems Engineering, Monash University,
  Melbourne, Australia. His research interests include computer
  vision, robotics, and 3-D video processing.
\end{IEEEbiographynophoto}%
\vspace*{-15mm}
\begin{IEEEbiographynophoto}{Y. Ahmet \c{S}ekercio\u{g}lu} is a
  professorial fellow at Heudiasyc Laboratory, Universite de
  Technologie de Compi{\`{e}}gne (UTC) in France. Between 2000 and
  2015, he was a member of the academic staff at the Department of
  Electrical and Computer Systems Engineering of Monash University,
  Melbourne, Australia. He established the Monash Wireless Sensor and
  Robot Networks Laboratory, and served as its director until his departure. He
  completed his Ph.D. degree at Swinburne University of Technology,
  Melbourne, Australia, and M.Sc. and B.Sc.  degrees at Middle East
  Technical University, Ankara, Turkey (all in Electrical and
  Electronics Engineering). He leads a number of research projects on
  distributed algorithms for self-organization in mobile visual
  sensor, ad hoc, and robot networks.
\end{IEEEbiographynophoto}%
\vspace*{-15mm}
\begin{IEEEbiographynophoto}{Tom Drummond}
  studied mathematics at Cambridge University for his first degree
  before immigrating to Australia in 1990. He worked for CSIRO in
  Melbourne until 1994 when he went to Perth to do his Ph.D. in
  Computer Science at Curtin University, Perth, Australia. He then
  returned to UK to undertake post-doctoral research in Computer
  Vision at Cambridge University and was appointed a permanent
  lectureship in 2002. In September 2010, he moved back to Melbourne
  and took up a professorship at Monash University, Melbourne,
  Australia.  His research interests include real-time computer
  vision, visually guided robotics, augmented reality, robust methods
  and SLAM.
\end{IEEEbiographynophoto}%
\vspace*{-12mm}
\begin{IEEEbiographynophoto}{Vincent Fr\'{e}mont}
  received the M.Sc. degree in Automatic Control and Computer Science
  from the Ecole Centrale de Nantes, France, in 2000 and the
  Ph.D. degree in Automatic Control and Computer Science from the
  Ecole Centrale de Nantes, France, in 2003.  He joined the Institut
  de Recherche en Communications et Cybernetique de Nantes in 2000,
  and the Laboratoire de Vision et Robotique de Bourges, in
  2003. Since 2005, he is an Associate Professor at the UTC. His
  research interests are computer vision for robotic perception with
  applications to intelligent vehicles.
\end{IEEEbiographynophoto}
\vspace*{-10mm}
\begin{IEEEbiographynophoto}{Enrico
    Natalizio}
  is an Associate Professor at the Universite de Technologie de
  Compi{\`{e}}gne (UTC), France, in the Network \& Optimization Group
  within the Heudiasyc Lab. He obtained his Ph.D.  from the Universita
  della Calabria, Italy and he was a visiting researcher at the
  Broadband Wireless Networking Laboratory at Georgia Tech in Atlanta,
  USA. Between 2005 and 2010, he worked as a research fellow and a
  contract professor at the Universita della Calabria, Italy. From
  2010 to 2012 he worked at INRIA Lille as a postdoctoral
  researcher. His current research interests include group
  communication in wireless robot and sensor networks and coordination
  and cooperation among swarm networked devices.
\end{IEEEbiographynophoto}%
\vspace*{-13mm}
\begin{IEEEbiographynophoto}{Isabelle Fantoni}
  graduated from the Universite de Technologie de Compi{\`{e}}gne
  (UTC), France, and received the Master of Sciences in Electronic
  System Design at the University of Cranfield in England (double
  degree), both in 1996. She received the Master in Control Systems,
  in 1997 and the Ph.D.  degree, in 2000 from the Universite de
  Technologie de Compi{\`{e}}gne (UTC). From 2001 to 2017, she was
  a permanent CNRS (French National Foundation for Scientific
  Research) Researcher at Heudiasyc Laboratory, UTC. Since September
  2017, she is with the Ecole Centrale de Nantes, CNRS, UMR 6004 LS2N,
  in Nantes.
\end{IEEEbiographynophoto}%
\end{document}